%% file: SPM_revpaper_main_final.tex
\PassOptionsToPackage{svgnames}{xcolor}

\documentclass[11pt,onecolumn,journal]{IEEEtran}

\usepackage[a4paper,top=2.5cm,bottom=2.5cm,left=3cm,right=3cm]{geometry}

\usepackage{amsmath,amsfonts}
\usepackage{algorithmic}
\usepackage{algorithm}
\usepackage{array}
\usepackage[caption=false,font=normalsize,labelfont=sf,textfont=sf]{subfig}
\usepackage{textcomp}
\usepackage{stfloats}
\usepackage{url}
\usepackage{verbatim}
\usepackage{graphicx}
\usepackage{cite}
\usepackage{hyperref}
\usepackage{booktabs}
\input{MathSymbolDefs2}

\usepackage{framed}
\usepackage[svgnames]{xcolor}
\definecolor{shadecolor}{rgb}{0.83, 0.83, 0.83}
\usepackage{setspace}

\begin{document}

\title{Low-Rank Tensor Decompositions for \\the Theory of Neural Networks}

\author{Ricardo Borsoi$^*$, Konstantin Usevich, Marianne Clausel
\thanks{R. Borsoi, K. Usevich and M. Clausel are with the Université de Lorraine, CNRS, CRAN, F-54000 Nancy, France. e-mail: {firstname.lastname}@univ-lorraine.fr. $^*$Corresponding author.}
}

\markboth{IEEE Signal Processing Magazine}%
{Borsoi \MakeLowercase{\textit{et al.}}: Low-rank Tensor Decompositions for the Theory of Neural Networks}

\maketitle

\begin{abstract}

The groundbreaking performance of deep neural networks (NNs) promoted a surge of interest in providing a mathematical basis to deep learning theory. Low-rank tensor decompositions are specially befitting for this task due to their close connection to NNs and their rich theoretical results. Different tensor decompositions have strong uniqueness guarantees, which allow for a direct interpretation of their factors, and polynomial time algorithms have been proposed to compute them. Through the connections between tensors and NNs, such results supported many important advances in the theory of NNs. In this review, we show how low-rank tensor methods---which have been a core tool in the signal processing and machine learning communities---play a fundamental role in theoretically explaining different aspects of the performance of deep NNs, including their expressivity, algorithmic learnability and computational hardness, generalization, and identifiability. Our goal is to give an accessible overview of existing approaches (developed by different communities, ranging from computer science to mathematics) in a coherent and unified way, and to open a broader perspective on the use of low-rank tensor decompositions for the theory of deep NNs.

\end{abstract}

\section{Introduction}
\label{sec:intro}

Deep learning achieved groundbreaking performance in a wide range of applications ranging from computer vision to natural language processing. 
In face of the practical success of deep neural networks (NNs) compared to the worst-case hardness found in some comparatively simple tasks (e.g., even learning 2-layer ReLU nets is NP-hard in network size), current algorithms, model architectures, etc. have converged to a set of practical frameworks that intrinsically exploit the structure of the problem to provide tractable solutions (such as in benign overfitting).

A good theoretical understanding of deep learning is essential for providing  proper design tools and further investigation directions. Thus, great effort has been dedicated to this problem by researchers from different communities, including computer science, mathematics, and signal processing, with important progress in understanding  expressivity, generalization, stability and implicit biases of NNs. In particular, mathematical tools that served as foundations for many algorithms in signal processing are being used to explain the underlying workings of deep learning, such as splines, kernel machines, sparsity, and, not the least, low-rank matrix and tensor decompositions, which are the focus of this paper.

In this review, we show how low-rank tensor methods~\cite{sidiropoulos2017tensor,comon2014spm,kolda2009tensor,hackbusch2012tensorBook,panagakis2021tensorsComputerVisionDeepLearning,anandkumar2014tensor,cichocki2017tensorNetworksPart1}---a fundamental mathematical object exploited for decades in the signal processing community and widely used in different applications involving deep NNs---play a fundamental role in theoretically explaining different aspects of the performance of deep learning. By shedding light on the various connections between tensors and NNs, a wealth of theoretical results about low-rank decompositions can be leveraged to study various behaviors of NNs, such as their expressivity/approximation, generalization and identifiability, as well as to develop learning algorithms with strong guarantees.
In this article, we aim at presenting  a unified perspective on a broad range of results on tensor decompositions for the theory of NNs, by covering the following topics:

\subsubsection{\textbf{NNs with low-rank weights}}
Many works represent weights in different NN architectures 
as low-rank tensors for their compression~\cite{novikov2015tensorizingNNs}, and also 
for the compression of their gradient updates (such as in LoRA~\cite{hu2022lora}). We will present results on the sensitivity and convergence %
of NN weight compression algorithms, on the generalization performance of such models, %
and on their implicit biases.
This will be covered in Section~\ref{sec:part1_lowRankWeights}.

\subsubsection{\textbf{Expressivity and approximation}}
Recent results draw on the intimate connection between tensors and specific classes of NNs---the \emph{sum-product networks} \cite{cohen2016tensorExpressivePowerNeuralNets}---to study the expressive power of different architectures (e.g., shallow vs. deep networks) \cite{cohen2016tensorExpressivePowerNeuralNets,lizaire2024expressivityRNNsTensor} and their function approximation properties in terms of smoothness (e.g., functions belonging to classical functional spaces as Besov) spaces \cite{ali2023approximationTreeTensorNetworks}. Other works used algebraic geometry to study the expressivity and identifiability of linear and polynomial NNs \cite{kileel2019expressivePolynomialNNs,usevich2025identifiabilityDeepPNNs,malgouyres2019multilinearCompressiveSensingLinearNNs}. This will be covered in Section~\ref{sec:part2_NNsAsTensorsSumProduct}.

\subsubsection{\textbf{Learning with differentiation}}
Through the connection between NNs and tensors obtained by \emph{derivatives},
the theory of low-rank decompositions helps to establish polynomial time training algorithms, generalization performance guarantees, handling trainable activation functions, and parameter identifiability for NNs. We will discuss results for learning 2- and 3-layer NNs that use this connection by leveraging different low-rank matrix and tensor formats \cite{janzamin2015beatingPerilsTensorNeuralNets,fornasier2021robustIdentificationShallowNeuralNets,dreesen2015decouplingPolinomials_1stOrder,ge2019learning2layerNNsSymmetricInputs}. This will be covered in Section~\ref{sec:part3_derivativesAndMoments}.

\subsubsection{\textbf{The use of tensors in emerging learning problems}}
We will consider the use of tensors to study the expressivity and learnability of generative models parametrized by polynomial NNs~\cite{chen2023learningPolynomialGenerativeModelsTensors}, and to study some classes of Hidden Markov Models (HMMs) and restricted Boltzmann machines (RBMs) \cite{glasser2019expressiveTensorNetworksProbabilisticModeling}. We will also review the use of tensors in parametrizing action-value functions in reinforcement learning (RL)~\cite{mahajan2021tesseract}, and in learning mixtures of linear classifiers \cite{chen2022mixtureLinearClassifiersTensor}.  This will be covered in Section~\ref{sec:part4_TensorsInotherLearningProblems}.

Sections~\ref{sec:tensorBackground}~and~\ref{sec:nnetBackground} contain background on tensor decompositions and neural networks, respectively, providing key definitions and references.

\section{Tensor decompositions definitions and background}
\label{sec:tensorBackground}

\subsection{\textbf{Notation}}

We mainly follow the notation of  \cite{comon2014spm}. Scalars, vectors and matrices are denoted by plain font ($x$ or $X$), lowercase bold font ($\bx$) and uppercase bold font ($\bX$), respectively. Tensors can be viewed as multidimensional arrays, and are represented by calligraphic font ($\tensor{X}$). The order of a tensor is the number of dimensions or \emph{modes}.
The notation $\tensor{X} \in \amsmathbb{R}^{I_1\times \cdots \times I_d}$ denotes an order-$d$ tensor of respective mode sizes.
The $(i,j,k)$-th element of $\tensor{X} \in \amsmathbb{R}^{I\times J \times K}$ is indexed as $\tensor{X}_{i,j,k}$;
``$:$'' denotes taking all elements in one mode (e.g., $\tensor{X}_{:,:,k}\in\amsmathbb{R}^{I\times J}$ is the $k$-th frontal slice of $\tensor{X}$). 
We use ``$\otimes$'' to denote tensor (outer) product between vectors,
so that $\ba\otimes\bb = \ba \bb^{\top}$ (a rank-one matrix) and $\tensor{X} = \ba\otimes\bb\otimes\bc$ denotes an order-3 tensor with $\tensor{X}_{ijk} = a_i b_j c_k$ (rank-1 order-3 tensor).
The tensor power of a vector $\ba$ is compactly written as
$\ba^{\otimes d} = \underbrace{\ba \otimes \cdots \otimes \ba}_{d \text{ times}}$.  
The inner product between two tensors is denoted $\langle\tensor{X},\tensor{Y}\rangle=\sum_{i_1,i_2,i_3} \tensor{X}_{i_1,i_2,i_3} \tensor{Y}_{i_1,i_2,i_3}$. 
Operators $\vspan(\cdot)$ and $\diag(\cdot)$ denote the vector space spanned by a set of vectors and a diagonal matrix, respectively. $\nabla^{(m)}$ denotes the $m$-th order differential operator.

\begin{figure*}
\begin{mdframed}[backgroundcolor=black!7]
    \centering
    \begin{minipage}{0.54\linewidth}
    \fontsize{9pt}{10.5pt}\selectfont
    \vspace{-0.3cm}
    The CPD of an order-3 tensor is a direct generalization of the matrix low-rank decomposition as a sum of rank-one components $\bX = \sum_{r=1}^R \ba_r \bb_r^{\top} = \bA \bB^{\top}$. 
    As with the matrix case, the vectors for different components can be grouped into \textit{factor matrices} (one for each mode).
    Inherent indeterminacies are present: permutation of rank-one components and rescaling of the vectors within each component (this corresponds to simultaneous permutations and rescaling of columns in the factor matrices).
    For order-$3$ tensors the CPD can be also rewritten as joint matrix factorization.
    \end{minipage}%
    \hfill
    \begin{minipage}{0.43\linewidth}
    \centering
    
    \includegraphics[width=\linewidth]{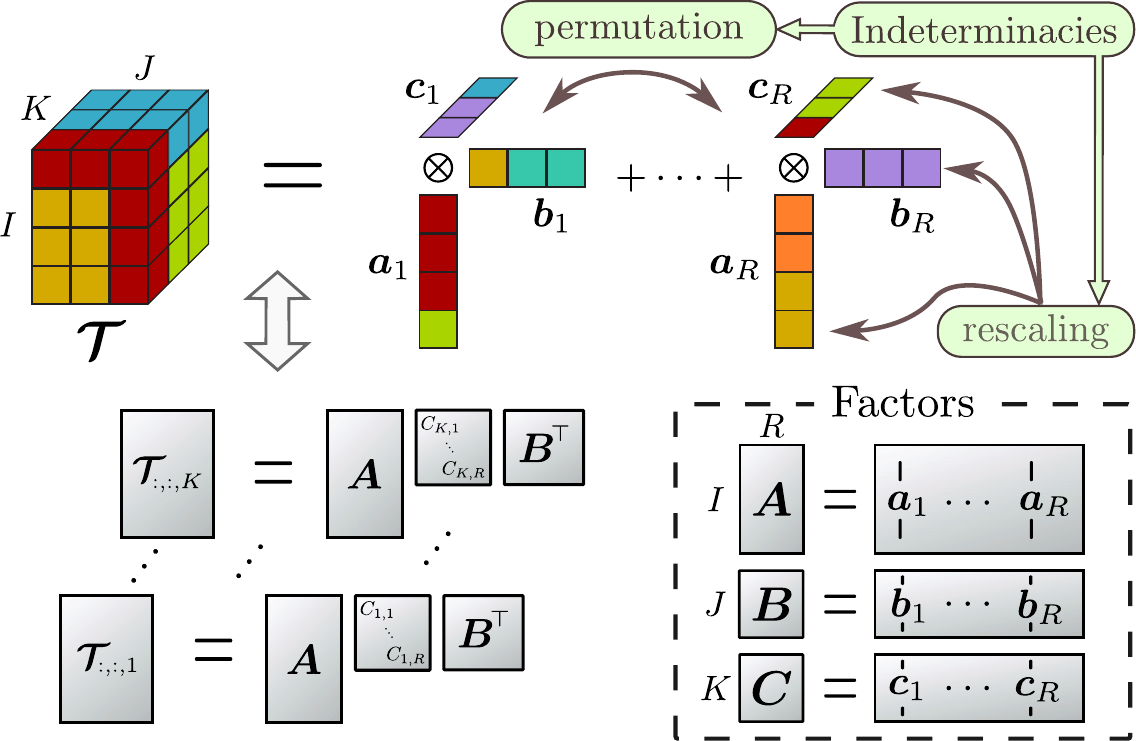} \hspace{-0.35cm}
    \end{minipage}
    \vspace{-0.2cm}
    \caption{The canonical polyadic decomposition.}
    \label{fig:CPD}
\end{mdframed}
\vspace{-1em}
\end{figure*}

\subsection{\textbf{Canonical polyadic decomposition (CPD)}}
\subsubsection{\textbf{Definition}}

The CPD is a natural generalization of  low-rank matrices, which  represents a tensor as a sum of $R$ rank-1 \emph{components}. For example,
an order-3 tensor $\tensor{T} \in \bbR^{I \times J \times K}$ admits an $R$-term CPD if 
\begin{equation}\label{eq:cpd}
    \tensor{T} = \sum\limits_{r=1}^{R} \ba_r \otimes \bb_r \otimes \bc_r\,, 
\end{equation}
where  $\ba_r \in \bbR^{I}$, $\bb_r \in \bbR^{J}$, and $\bc_r \in \bbR^{K}$ are   the \emph{factors}   of the CPD. 
The number of components $R$ is called the \textit{tensor rank} (or CP rank) if it is minimal. Scalar weights for components are often added into definition of the CPD \eqref{eq:cpd} (e.g.,  $\tensor{T} = \sum\limits_{r=1}^{R} \lambda_r \ba_r \otimes \bb_r \otimes \bc_r$), but we omit them for simplicity.  
The CPD can be also explicitly written in terms of \emph{factor matrices} as $\tensor{T}_{ijk} = \sum\limits_{r=1}^{R} A_{ir} B_{jr} C_{kr}$,
where
 $\bA \in \bbR^{I \times R}$, $\bB \in \bbR^{I \times R}$ and $\bC \in \bbR^{K \times R}$ collect the factors as their columns (see Fig.~\ref{fig:CPD} for an illustration).
Also, for order-$3$ tensors, the   CPD has an interpretation  as joint matrix factorizations with shared factors: each $k$-th frontal slice of $\tensor{T}$ can be written as $\tensor{T}_{:,:,k} = \bA \diag(C_{k,:}) \bB^\top$.
Unlike the (single) matrix case, the rank $R$ of the CPD can exceed its dimensions (e.g., $\bA$ can have more columns than  rows).

\subsubsection{\textbf{Uniqueness}} 
An important feature of the CPD \eqref{eq:cpd} of tensors of order $\ge 3$ is that the decomposition is unique (under some mild conditions) up to trivial indeterminacies (e.g., permuting rank-one components, or rescaling the factors within each component), as illustrated in Fig.~\ref{fig:CPD}. On the other hand, uniqueness does not happen for matrices (order-$2$ tensors) as in $\bX = \bA\bB^{\top}$, since $\bB$ can come from any basis.
The earliest conditions for CPD uniqueness are due to Kruskal (which in the simplest case require $\bA$ and $\bB$ full column rank and $\bC$ not having proportional columns). More recently, these results were significantly improved (see e.g.,  \cite{sidiropoulos2017tensor,oneto2025ranks} and references therein), and the CPD can be unique for a very high number of components exceeding the dimensions (for example, up to roughly $\frac{I^2}{3}$ for cubic $I \times I \times I$ tensors).
The uniqueness properties have a direct implication  for the interpretability and explainability of NNs, as it will be shown later.

\vspace{-0.5em}

\subsection{\textbf{Tucker (multilinear) decomposition and tensor networks decompositions}}

\subsubsection{\textbf{Tucker decomposition}}

The second basic tensor decomposition is the Tucker (or multilinear) decomposition \cite{kolda2009tensor}.
A Tucker decomposition  of a third-order tensor $\tensor{T}$ is of the following form: %
\begin{align}\label{eq:tucker}
\tensor{T}_{ijk} = \sum\limits_{r_1=1}^{R_1} \sum\limits_{r_2=1}^{R_2} \sum\limits_{r_3=1}^{R_3} \tensor{G}_{r_1,r_2,r_3} U_{i,r_1} V_{j,r_2} W_{k,r_3} \,,
\end{align}
where %
$\tensor{G}\in\amsmathbb{R}^{R_1\times R_2\times R_3}$ is the core tensor and $\bU \in \bbR^{I \times R_1}$, $\bV \in \bbR^{J \times R_2}$, $\bW \in \bbR^{K \times R_3}$.
The Tucker decomposition is linked to matrix ranks of the matricizations (called \textit{multilinear ranks}, and written as the tuple $(R_1,R_2,R_3)$); this  implies that the Tucker decomposition exists for some $R_1 \le I$, $R_2 \le J$, $R_1 \le K$.
The multilinear ranks are in general different from the CP rank.
In fact, the CPD can be seen as a Tucker format with a very structured (diagonal) core tensor $\tensor{G}$, 
which makes the Tucker decomposition more flexible.
However, unlike the CPD, the Tucker decomposition does not have uniqueness properties.

\begin{figure*}
\begin{mdframed}[backgroundcolor=black!7]
    \centering
    \begin{minipage}{0.41\linewidth}
    \fontsize{9pt}{10.5pt}\selectfont

 \vspace{-0.25cm}
    
    Tensor tree network formats (left) and the corresponding trees (right).
     The nodes in a tree correspond to factors, whose  orders are equal to the number of edges adjacent to a node.
     Factor dimensions are given by \emph{bond dimensions} specified for each edge.
     There are two types of edges: ordinary and dangling.
     Summation is performed along all ordinary edges and the dangling edges define the output dimensions of a tensor.
    \end{minipage}%
    \hfill
    \begin{minipage}{0.58\linewidth}
    \centering
    \includegraphics[width=0.95\linewidth]{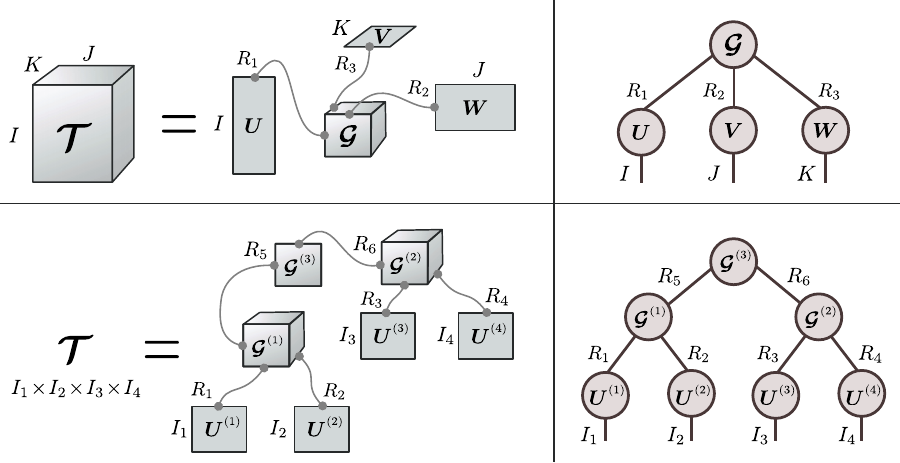} \hspace{-0.5cm}
    \end{minipage}
    \vspace{-0.1cm}
    \caption{Tucker and tensor tree networks.}
    \label{fig:tensor_formats}
\end{mdframed}
\vspace{-1em}
\end{figure*}

\subsubsection{\textbf{Tensor networks and tensor trains}}
A tree-based tensor network decomposition \cite{hackbusch2012tensorBook} is a generalization of the Tucker decomposition 
defined by a tree, where the nodes correspond to factors and summation is performed along non-dangling edges.
The Tucker decomposition is the simplest example of tensor network (see Fig.~\ref{fig:tensor_formats}, top row).
The second example of the tree in Fig.~\ref{fig:tensor_formats} gives rise to
\begin{equation}\label{eq:HT}
\tensor{T}_{i_1,i_2,i_2,i_3} =
\sum\limits_{r_5=1}^{R_5} \sum\limits_{r_6=1}^{R_6}
\tensor{G}^{(3)}_{r_5,r_6}
\left(\sum\limits_{r_1=1}^{R_1} \sum\limits_{r_2=1}^{R_2}
\tensor{G}^{(1)}_{r_5,r_1,r_2} 
U^{(1)}_{i_1,r_1}
U^{(2)}_{i_2,r_2}
\right)
\left(\sum\limits_{r_3=1}^{R_3} \sum\limits_{r_4=1}^{R_4}
\tensor{G}^{(2)}_{r_6,r_3,r_4}
U^{(3)}_{i_3,r_3}
U^{(4)}_{i_4,r_4}\right);
\end{equation}
such a decomposition is the case of the popular hierarchical Tucker (HT) format, which is typically defined for a binary tree.
Another popular tensor format which proved to be very efficient for machine learning tasks is the tensor train (TT), which corresponds to the minimal (linear) tree.

The key advantage of tree tensor formats over Tucker is the storage complexity: the size of Tucker core grows exponentially with the tensor order, while in TT or HT the factors are usually at most order-3 tensors. Thus, tree tensor formats achieve significant data compression and have been widely used. 
However, as with the Tucker decomposition, tree tensor decompositions are not unique.

\vspace{-0.25em}

\subsection{\textbf{Key properties of tensor decompositions}}
\subsubsection{\textbf{Generic properties}} 
A wealth of results exist on \emph{generic properties} of tensor decompositions, that is, the properties which hold for almost all tensors (as explained below) of a given size or tensor format. Generic properties can be studied using the powerful tools of algebraic geometry \cite{oneto2025ranks}.
A key example is the \emph{generic tensor rank}:
all complex-valued tensors of given size $I_1\times \cdots \times I_d$ have the same CP rank $\genrank\triangleq \genrank(I_1,\ldots,I_d)$, except for a set of Lebesgue measure zero.
In other words, if a complex tensor of a given size is drawn according to continuous probability distribution, then with probability one the tensor has CP rank $\genrank$.
For example, for cubic tensors $\genrank(I,I,I) = \big\lceil\frac{I^3}{3I-2}\big\rceil$, and for many different tensor shapes the generic ranks are known. The generic ranks also give a bound $\maxrank \le 2\genrank$ on maximal rank $\maxrank$ of a given tensor size, which is otherwise difficult to study and unknown for most cases.
For real-valued tensors, several \emph{typical ranks} can occur (corresponding to sets of positive Lebesque measure), but the smallest typical rank is equal to $\genrank$, see \cite{comon2014spm,comon2017identifiabilityXrank,oneto2025ranks} for an overview.

For other tensor formats with multiple ranks (e.g., $(R_1,\ldots,R_d)$ for HT or TT), one studies the  generic properties of tensors of bounded ranks.
For example, for a $d$-th order generic tensor with all TT/HT ranks bounded by $R_k<r$, $k=1,\ldots,d$ (i.e., for all tensors of bounded ranks except for a set of factors of Lebesgue measure zero), its CP rank is at least $r^{\frac{d}{2}}$ \cite{khrulkov2018expressiveTensorNNs}. 
Maximal and generic ranks are fundamental in the study of the expressivity of a NN, and can compare the expressive power of different architectures (e.g., shallow vs. deep, or those corresponding to different ranks of some coefficient tensor).

Finally, another key property is the \textit{generic uniqueness} of tensor decompositions. For tensors of given size (and bounded tensor rank), their decomposition of rank-$r$ is generically unique if it is unique except for a set of factors of Lebesgue measure zero (that is, the tensor space is $r$-identifiable). 
Generic uniqueness cannot happens for $r > \genrank$ and rarely occurs for $r= \genrank$. Recent results show that identifiability holds for most of the ranks below $\genrank$, thanks to recent breakthroughs in algebraic geometry \cite{oneto2025ranks}.
For example, for cubic $I \times I\times I$ tensors, the CPD is generically unique ($r$-identifiable) for $r < \big\lceil\frac{I^3}{3I-2}\big\rceil -1$ \cite{oneto2025ranks}.

\subsubsection{\textbf{Computing the decompositions and their approximations}}
Although the CPD is parameter-efficient (using $O\big(R(I+J+K)\big)$ parameters to represent $\tensor{T}$),
its computation is an ill-posed problem and generally NP-hard. Nonetheless, there are polynomial time (and even algebraic) algorithms to compute the CPD under some conditions \cite{sidiropoulos2017tensor}, mainly based on reducing it to eigenvalue computations.
In practice, local optimization schemes (such as block-coordinate descent) are often used and show good performance.

Understanding the \emph{robustness of the CPD to perturbations} is of prime importance for computing CPD approximations, and several advances have been made recently, such as    %
perturbation analysis \cite{evert2022lra} and random tensor models \cite{arous2019landscape}.
There exist stronger results on decomposability of perturbed tensors in polynomial time using power iteration~\cite{anandkumar2014tensor}, but for  particular constraints on the factors (near-orthogonality). Such results are intimately linked to the (in)existence of polynomial time algorithms to learn NNs~\cite{mondelli2019connection2layerNnetTensorDec}.

Unlike the CPD,  computing approximations in Tucker, TT and HT formats is a tractable and well-posed problem. Such tensor formats enjoy efficient and stable algorithms based on linear algebra tools such as the SVD \cite{cichocki2017tensorNetworksPart1,sidiropoulos2017tensor}.
The approximation properties for Tucker-based decompositions are also well-studied \cite{hackbusch2012tensorBook,ali2023approximationTreeTensorNetworks}, and allows for controlling the error when approximating some tensor in this format.

\subsubsection{\textbf{Other factorizations}}
Despite being widely used to explain the behavior of NNs, the CPD has limited flexibility, which is one reason many results only address the 2-layer case (see, e.g.,~\cite{janzamin2015beatingPerilsTensorNeuralNets,mondelli2019connection2layerNnetTensorDec}).
Different generalizations of the CPD exist: symmetric, partially symmetric (where symmetry is imposed on factors of the rank-one components), block-term decomposition (BTD, decomposition of a tensor as a sum of low-rank tensors); and the
so-called tubal tensor rank decomposition, or t-SVD (often used for multichannel/3D data) which is a special case of the BTD with a fixed factor matrix \cite{gilman2022tsvd}.
The above decompositions are additive, and most of them fall  under the umbrella of X-rank decompositions~\cite{comon2017identifiabilityXrank,oneto2025ranks}.
Many generic properties of such decompositions (e.g., identifiability) can be studied using the tools from {algebraic} geometry, similarly to the case of CPD.
Besides additive decompositions, there are other  generalizations of the CPD such as the Paratuck-2~\cite{kolda2009tensor}, which enjoys uniqueness guarantees and can be useful in the analysis of deep learning models.%

\section{Neural networks: basic setup}
\label{sec:nnetBackground}

Mathematically, a feedforward NN with $L$ layers is a function $f:\mathscr{X}\to \mathscr{Y}$ with domain $\mathscr{X}\subset\bbR^\indim$ and codomain $\mathscr{Y}\subset\bbR^\outdim$ consisting on a layerwise transformation parameterized as 
\begin{equation}\label{eq:ffNN}
\by \,=\, f(\bx)
\,\triangleq\, f_L(f_{L-1}(\ldots f_1(\bx)\ldots )) \,,
\end{equation}
where each function $f_{\ell}:\bh_{\ell-1}\mapsto\bh_{\ell}$ represents the transformation taking place in the $\ell$-th layer, which consists on the composition of an affine transformation and a (typically elementwise) nonlinearity:
\[
\bh_{\ell} \,=\, f_{\ell}(\bh_{\ell-1})
    \,\triangleq\, \sigma_{\ell}(\bA_{\ell} \bh_{\ell-1} + \bb_{\ell}) \,, \qquad  \text{(hidden layers)}
\]
where $\bh_{\ell}\in\bbR^{d_{\ell}}$ denotes the \emph{hidden representation} at layer $\ell\in\{1,\ldots,L-1\}$, with $d_{\ell}$ neurons. For the first and last layers ($\ell=0$ and $\ell=L$), $\bh_{0}\triangleq \bx$ and $\bh_{L}\triangleq \by$ are defined to be the NN input and output, respectively. For each hidden layer, $\bA_{\ell}\in\amsmathbb{R}^{d_{\ell}\times d_{\ell-1}}$ is a linear transformation and $\bb_{\ell}\in\amsmathbb{R}^{d_{\ell}}$ is a bias term.
While this presentation focuses on multilayer perceptrons (MLPs) for the sake of simplicity, low-rank tensor decompositions have had significant practical impact in various other architectures (especially in reducing their computation and memory footprint), including transformers, convolutional NNs (CNNs) and recurrent NNs (RNNs) \cite{liu2023tensorCompressionNNsReview}, which are widely used in computer vision and sequence modeling.

Given some probability measure $p(\bx,\by)$ on $\mathscr{X}\times \mathscr{Y}$, we focus on the supervised learning task, whose aim is to learn the parameters $\{\bA_{\ell},\bb_{\ell}\}_{\ell=1}^L$ of a NN so as to minimize the expectation of some risk function $\riskloss:\mathscr{Y}\times \mathscr{Y}\to\amsmathbb{R}$:
\begin{align}
    \nonumber
    \mathscr{R}(f) = \Ex_{p(\bx,\by)}\{\riskloss(\by,f(\bx))\} \,,
\end{align}
where $\Ex\{\cdot\}$ denotes the expectation operator. In practice, the probability measure $p(\bx,\by)$ is typically unknown and only a dataset of i.i.d. samples $(\bx_n,\by_n)\sim p(\bx,\by)$, $n=1,\ldots,N$ is available. This leads to the %
minimization of the empirical risk $\widehat{\mathscr{R}}(f) = \frac{1}{N} \sum_{n=1}^N \riskloss(\by_n,f(\bx_n))$. Other tasks such as generative modeling and reinforcement learning follow a different setup; they will be discussed later in the paper.

\begin{table}[t]
    \centering    
    \scriptsize
    \caption{Low-rank tensor formats and their use in the theory of NNs.}
    \vspace{-0.7em}
    \resizebox{\linewidth}{!}{%
    \begin{tabular}{c|cccccccc}
    \hline
    Tensor format & \makecell{Compression \\ capability} & Uniqueness  & \makecell{Computing \\ tractability} & \makecell{Key features and \\ use in NNs} \\ \hline
    \arrayrulecolor[gray]{0.85} 
    
    CP      &  $\bullet\bullet$  & \newcheckmark & \makecell{NP-hard in general \\ but well studied} & \makecell{Known generic ranks (linked to expressivity of NNs); \\ NN learning with derivatives and flexible activations} \\[0.2cm]\hline
    
    Symmetric CP & $\bullet$  & \newcheckmark & \makecell{Good polynomial time algorithms \\ for some rank values} & \makecell{Stable under perturbations; key tool for \\ learning NNs with derivatives/method of moments} \\[0.2cm]\hline
    
    Tucker & $\bullet\bullet\bullet$ & \newcrossmark & Efficient algorithms based on linear algebra & \makecell{Widely used for compression of NN weights} \\[0.2cm]\hline
    
    TT and HT & $\bullet\bullet\bullet$  & \newcrossmark & Efficient algorithms based on linear algebra  & \makecell{Known generic ranks (linked to expressivity of NNs);\\ well-adapted to work in high-dimensions; directly \\ linked to (the study of) sum-product architectures} \\[0.2cm]\hline
    
    Paratuck &  $\bullet\bullet$  & \newcheckmark & \makecell{Hard to compute} & Linked to deep ($>2$ layers) NNs by derivatives \\ \arrayrulecolor{black} \hline
    \end{tabular}}
    \label{tab:overview_tensor_uses_NNs}
    \vspace{-1em}
\end{table}

The algorithms that have been proposed to learn NNs and the different types of architectures can be quite diverse. Nonetheless, some theoretical questions are relevant to most existing algorithms:

\begin{itemize}
    \item \textbf{\textit{Expressivity and approximation}}: what classes of functions can a given NN architecture $f(\bx)$ represent? How can we compare the capacity of different NNs, including shallow vs. deep or full vs. compressed models? How well can it approximate smooth functions (e.g., in Besov spaces)?
    
    \item What is the \textbf{\textit{generalization performance}} of a given NN? That is, for an NN ${f}$ and a distribution $p(\bx,\by)$, can we upper bound its expected risk $\mathscr{R}({f})$ as a function of its empirical version $\widehat{\mathscr{R}}({f})$? What is the impact of the architecture and learning method on generalization?
    
    \item \textbf{\textit{Identifiability}}: when are the NN weights $\{\bA_{\ell},\bb_{\ell}\}_{\ell=1}^{L}$ uniquely recovered (up to trivial ambiguities, such as permutations)? This key property supports the interpretability and explainability of NNs.
    
    \item \textbf{\textit{Learnability}}: when can a NN be learned in polynomial time? For which architectures, and by which classes of algorithms?
\end{itemize}

We aim to provide insight into these questions by leveraging the connections between NNs and tensor decompositions. An overview of the different tensor formats and their use in the study of NNs is provided in Table~\ref{tab:overview_tensor_uses_NNs}. 
In particular, the following sections will review: 
1) the use of tensors to compress NN weights and its impact on performance (Section \ref{sec:part1_lowRankWeights});
2) specific NN architectures directly connected to tensors and their expressivity (Section~\ref{sec:part2_NNsAsTensorsSumProduct});
3) the use of differentiation and the method of moments to learn NNs (Section~\ref{sec:part3_derivativesAndMoments}); and
4) the emerging use of tensors in generative modeling, RL and learning mixtures of linear classifiers (Section~\ref{sec:part4_TensorsInotherLearningProblems}).

\section{Low-rank parametrization of NN weights}
\label{sec:part1_lowRankWeights}

\begin{singlespace}
\begin{mytcolorbox}[title=\textbf{Summary}: NNs whose weights are parametrized as low-rank tensors for compression purposes.]
\begin{itemize}
    \item[+] Compression of NN weights $\bA_{\ell},\bb_{\ell}$ (or gradients, during finetuning) which are represented as low-rank tensors, reduces storage cost and inference time. %
    \item[+] Widely used, applied to many architectures and problem settings.
    \item[-] The theoretical impact of low-rank formats on NN generalization and training, as well as implicit biases towards low-rank formats is hard to study due to its interaction with the high nonlinearity in common NN architectures.
\end{itemize}

\tcbsubtitle{\textbf{Key results leveraged from low-rank tensor decompositions:}}
\begin{itemize}
    \item Capability to represent matrices and tensors with a reduced number of parameters (essential for compression).
    \item Efficient tools for optimization in low-rank formats.
    \item \textbf{Commonly used tensor formats}: CPD, Tucker, TT, among others.
\end{itemize}
\end{mytcolorbox}
\end{singlespace}

\vspace{-0.5em}

\subsection{\textbf{Compression of weights}}

\begin{figure*}
\begin{mdframed}[backgroundcolor=black!7]
    \centering
    \begin{minipage}{0.39\linewidth}
    \fontsize{9pt}{10.5pt}\selectfont
    The \textbf{parametrization of NN weights as low-rank tensors} proved to be a powerful technique for the compression of large models, reducing the costs involved with their training and inference. By reducing the amount of model parameters, low-rank representations reduce the flexibility of the function class, which, despite reducing its approximation capacity, can positively impact the generalization of the model.
    \end{minipage}%
    \hfill
    \begin{minipage}{0.5999\linewidth}
    \centering
    \includegraphics[width=0.9\linewidth]{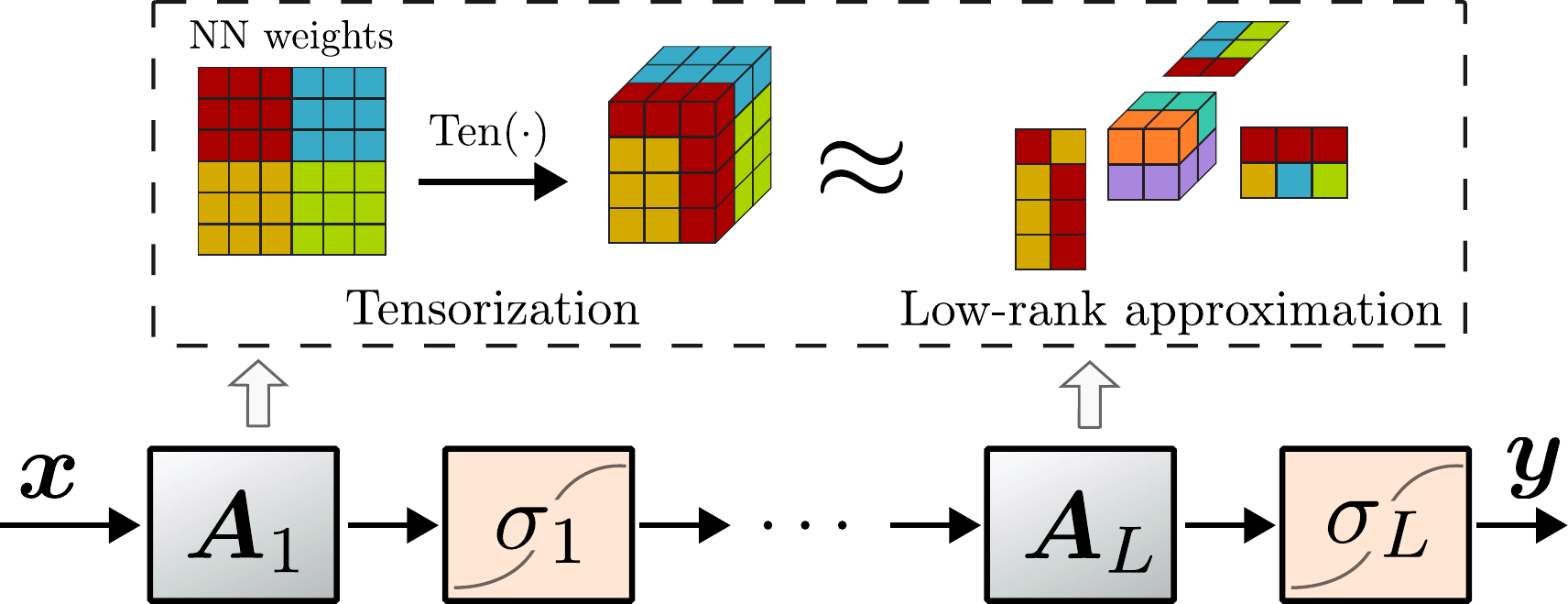} \hspace{-0.35cm}
    \end{minipage}
    \vspace{0.2cm}
    \caption{Illustration of the use of low-rank tensor formats for the compression of tensorized NN weight matrices.}
    \label{fig:figure1_tensorization_weight_compression}
\end{mdframed}
\vspace{-0.5em}
\end{figure*}

Low-rank tensor decompositions have become a popular approach for the compression of NN weights \cite{liu2023tensorCompressionNNsReview}. In particular, tensorial approaches are an appealing alternative that complements and can be naturally combined with classical compression methods such as quantization, pruning and knowledge distillation since they can exploit the natural low-rank representations which appears when NN weights are highly correlated. %
When confronted with NNs whose weights $\bA_{\ell}$ are very large, they can provide a low-dimensional parametrization of $\bA_{\ell}$ that drastically reduces the computation and storage costs of NN learning and inference.
Earliest uses of tensor approximations appear for CNNs, where the layers are parameterized by convolution kernel tensors $\tensor{A}_{\ell}\in\amsmathbb{R}^{c_1 \times c_2 \times \kappa_h \times \kappa_w}$ (\# of input channels $\times$ \# of output channels $\times$ kernel height $\times$ kernel width), which can be directly compressed using a low-rank format. 
However, the role of tensor decomposition in NN compression is much more broad. For MLPs, a key point is to define a map $\tensorize(\cdot)$ that orders a matrix $\bA_{\ell}$ in the form of a tensor:
\begin{align}
    \tensorize(\bA_{\ell}) = \tensor{W}_{\ell} \in \amsmathbb{R}^{n_1 \times \cdots \times n_K} \,.
\end{align}
Tensor decompositions can then be used to represent $\tensor{W}_{\ell}$ with low rank~\cite{novikov2015tensorizingNNs}. 
This approach, depicted in Fig.~\ref{fig:figure1_tensorization_weight_compression}, is called the \emph{tensorization} of linear layers. The use of tensor decompositions for compression is not limited to MLPs: various NN architectures including CNNs, RNNs and transformers have been successfully compressed using different low-rank formats (CP, Tucker, TT, etc.).
See \cite{liu2023tensorCompressionNNsReview} and references therein for an overview of low-rank compression choices.

Note that tensorization has been used not only to compress NNs but also to increase their expressiveness, as a means of capturing higher order interactions between inputs/hidden variables. This was used in tensor attention modules in transformers %
and in second-order RNNs~\cite{lizaire2024expressivityRNNsTensor}, which were proven to have increased expressiveness over their original counterparts. 
However, the most common use of low-rank representations is still to reduce not only storage costs but also training and inference time. This is because in many cases the computations can be performed using the decomposed format itself, without need to reconstruct the full weight tensor (which would not decrease the computation time during inference).

The availability of different low-rank formats introduces the question of what is the best loss function or tensor rank for compressing a given NN architecture. Relevant work in this direction include using stable loss functions to avoid degenerate components when compressing CNN layers (see \cite{panagakis2021tensorsComputerVisionDeepLearning}), and \cite{zangrando2024geometryAwareTrainingTucker}, which uses the Riemannian geometry of the Tucker decomposition to train NNs in compressed format with improved convergence performance, besides proposing an adaptive rank selection strategy.

\subsubsection{\textbf{Low-rank updates (compression of gradients)}} 
Besides  compression of weights, there has been a recent explosion in the popularity of low-rank adaptation (LoRA) techniques \cite{hu2022lora}, where low-rank approximations of gradient updates of weights are used for fine tuning transformer architectures (e.g., in LLMs) in environments with limited data and/or computation resources.
In a nutshell, instead of minimizing some loss $\mathscr{L}(\mathbf{A})$ with respect to a full weight matrix $\mathbf{A}$, LoRA proposes to minimize the loss $\mathscr{L}(\mathbf{A}_{\rm pre} + \mathbf{U}\mathbf{V}^{\top})$, where $\mathbf{A}_{\rm pre}$ is a pretrained weight matrix and $ \mathbf{U}\mathbf{V}^{\top}$ is a low-rank update.
Recent work has extended LoRA to a tensorial compression of weights, which can further reduce the number of parameters in the  updates~\cite{yang2024loretta_lowRankTensorAdaptation}. However, such approaches still lack detailed theoretical studies. %

\vspace{-0.5em}

\subsection{\textbf{Theoretical results}}

Despite the practical impact of low-rank compression approaches in reducing the computation and storage costs, few theoretical results are available. The difficulties come from the fact that the low-rank models (which are, by themselves, nonlinear) appear inside highly nonlinear network architectures. Nonetheless, recent effort has made important progress in the study of both generalization and inductive bias of NNs with low-rank weights in tensor formats.

\subsubsection{\textbf{Generalization}}
Recent works studied the generalization performance of NNs with layers represented in tensor formats by considering classical (e.g., covering number, Rademacher) measures of complexity.
The compression of convolutional layers of a pretrained CNN $f$ using a CPD with ranks $\{R_{\ell}\}_{\ell=1}^L$ was considered in~\cite{li2020generalizationNNsTensorLayersCompression}. The authors derive bounds for the generalization performance of the compressed network $f_{\rm comp}$.
First, it is shown that for well-chosen CP ranks the compressed CNN will be close to the original one, that is, $\|f(\bx)-f_{\rm comp}(\bx)\|\leq\epsilon\|f(\bx)\|$. Using this result, the expected risk of the compressed CNN can then be bounded as $\mathscr{R}(f_{\rm comp})-\widehat{\mathscr{R}}(f)\leq O\big(\sqrt{\sum_{\ell=1}^{L} R_{\ell} (c_1+c_2+\kappa^2)}\big)$, where $c_1$, $c_2$, and $\kappa$ are the number of input/output channels and the kernel dimensions, respectively. This sheds light on how the choice of rank impacts the NN generalization.

Another work \cite{wang2024lowrankNN_tSVD_robustgeneralizationbias} considered NNs with linear layers parametrized with the {tubal tensor format} (\emph{t-NN layers}). Upper bounds were derived for the robust (i.e., under adversarial attacks) generalization error of such NNs by extending bounds on the Rademacher complexity of regular NNs to the class of networks $f_{\rm tNN}$ with t-NN layers of ranks $\{R_{\ell}\}_{\ell=1}^L$, scaling as $\mathscr{R}(f_{\rm tNN})-\widehat{\mathscr{R}}(f_{\rm tNN})\leq O\big(\sqrt{\sum_{\ell=1}^{L} R_{\ell} (d_{\ell}+d_{\ell-1})}\big)$. Thus, although decreasing its flexibility, low-rank weights can lead to better generalization in this class of NNs.

\subsubsection{\textbf{Neural collapse}}
Many recent works studied the so-called \emph{neural collapse}, an implicit bias of NN training algorithms (especially gradient descent) towards solutions with low-rank weights~\cite{papyan2020neuralCollapseNNpnas}. 
More precisely, neural collapse was first observed in a classification context in the last layer of MLPs, and was later noticed to occur in intermediate layers which also converge towards low-rank matrices. This behavior occurs during the so-called \textit{terminal phase} of training, that is, a phase beyond the epoch where the training error vanishes and the norm of the gradient is very small. Thus, neural collapse tends to be theoretically studied for gradient descent under very small stepsizes.

While such inductive biases in training MLPs with matrix formats~\cite{papyan2020neuralCollapseNNpnas}
have been well-investigated, the study of nonlinear NNs with weights in tensor formats has only started more recently.
In \cite{wang2024lowrankNN_tSVD_robustgeneralizationbias}, the authors showed that gradient flow (gradient descent under very small stepsizes) on NNs with linear layers in the tubal tensor format (t-NN layers) is biased towards low-rank solutions, demonstrating that neural collapse can also occur in the tensorial case.
Note, however, that neural collapse highly depends on the loss function and optimization strategy used in training.

\section{Neural network architectures as expressive tensor formats}
\label{sec:part2_NNsAsTensorsSumProduct}

\begin{singlespace}
\begin{mytcolorbox}[title=\textbf{Summary}: NNs architectures directly linked to a tensor format (sum-product and polynomial NNs).]

\begin{itemize}
    \item[+] Restricting to architectures more closely linked to tensor formats (e.g., sum-product nets, which are linked to CPD, TT, etc.) allow many powerful results to be leveraged.
    \item[+] Results on the expressivity of tensorized NNs can be linked to the rank of a  tensor in a low-rank format; this can be used to compare  architectures (e.g., shallow vs. deep, or using different ranks).
    \item[+] Parameter ideitifiability results for linear and some polynomial NNs.
    \item[-] The results do not necessarily generalize to other more common NN architectures. 
\end{itemize}

\tcbsubtitle{\textbf{Key results leveraged from low-rank tensor decompositions:}}
\begin{itemize}
    \item Generic ranks of tensors, relation between ranks of different formats (e.g., CP vs. TT).
    \item Dimension of algebraic varieties.
    \item \textbf{Commonly used tensor formats}: TT, HT.
\end{itemize}
\end{mytcolorbox}
\end{singlespace}

Studying general NNs models without further assumptions can be intractable. However, particular architectures, such as sum-product and polynomial networks, have been investigated in more depth. Such networks are highly expressive, and their close connection to well-mastered tensor formats (including the CPD, TT, HT and X-rank decompositions) provided support to powerful theoretical results on the expressivity, approximation, and weight identification of such networks.
Although many real-world applications currently use overparametrized NNs that naturally lead to highly expressive models, theoretically studying the expressivity of different architectures can shed light on how specific choices (e.g., number of layers) impact the class of functions the NN can learn. This elucidates fundamental questions such as width \textit{versus} depth trade-offs. Moreover, complementary questions such as which small NN architectures can represent a desired class of functions, or that are also identifiable and might have better interpretability properties are of significant practical interest.
In the following, we review existing results, including linear, sum-product, and polynomial NNs.

\subsection{\textbf{Common architectures: stability and identification}}

\subsubsection{\textbf{Linear networks}}
Although linear NNs are generally not identifiable, considering sparse or Toeplitz-structured weight matrices provides a setting where fine-grained results about weight identification and stability can be derived~\cite{malgouyres2019multilinearCompressiveSensingLinearNNs}. Such problems can be studied through the lens of \textit{linearly parametrized networks} of the form $f(\bx)=\bA_L(\bh_L)\cdots\bA_1(\bh_1)\bx$, where $\bA_{\ell}(\bh_{\ell})$ are structured weight matrices depending on low-dimensional vectors of parameters $\bh_{\ell}\in\amsmathbb{R}^H$. A key idea is the use of the so-called \emph{tensorial lifting} approach, where the linear function $f(\cdot)=\bA_L(\bh_L)\cdots\bA_1(\bh_1)$ can be represented by a highly structured linear operator $\mathscr{M}$ applied to the order-$L$ rank-1 tensor $\bh_1\otimes \cdots\otimes \bh_L$ as
\begin{align}
    f(\cdot) = \mathscr{M}\big(\bh_1\otimes \cdots\otimes \bh_L\big) \in \amsmathbb{R}^{\outdim \times \indim} \,.
\end{align}
Using algebro-geometrical properties of the space of rank-1 and rank-2 tensors and the nullspace of operator $\mathscr{M}$ (as well as its conditioning), identifiability and stability results for the weights were obtained, meaning that if two structured networks $f$ and $\hat{f}$ with parameters $\{\bh_\ell\}_{\ell}$ and $\{\hat{\bh}_\ell\}_{\ell}$ have similar outputs, i.e., $f(\bx)\approx \hat{f}(\bx)$, then their parameters will also be similar, up to trivial ambiguities.

\subsubsection{\textbf{Polynomial neural networks}}
An important class of NNs are \emph{polynomial networks} (PNNs) \cite{kileel2019expressivePolynomialNNs}, that is, NNs $f(\bx)$ whose  activations are polynomials of fixed degree. 
The function space they generate forms an \emph{algebraic variety} of polynomials, allowing the use of tools from algebraic geometry to investigate its properties. 
The basic case uses \emph{monomial activations}, where $\sigma_{\ell}(\cdot) = (\cdot)^K$ with fixed $K$ ($K$-th degree power of inputs).
By convention, we assume there is no nonlinear activation at the last layer.

To explain the connection between PNNs and tensors, let us consider the case of monomial activations and no biases studied in in \cite{kileel2019expressivePolynomialNNs} (although biases can also be easily treated \cite{usevich2025identifiabilityDeepPNNs}).
In such a case, the NN is a \emph{homogeneous polynomial} of degree $K^{L-1}$ in $\bx = (x_1,\ldots,x_{d_{in}})$, i.e., only monomials of the same power appear.
Each output $f_i(\bx)$ then corresponds to a different symmetric tensor $\tensor{F}^{(i)}$ of order $K^{L-1}$:
\[
   f_i(\bx)= \big\langle\tensor{F}^{(i)}, \underbrace{\bx \otimes \cdots \otimes \bx}_{K^{L-1} \text{ times}}\big\rangle \,.
\]
For example, for quadratic polynomials over $\bx=(x_1, x_2)$, we have $a x_1^2 + 2b x_1 x_2+c x_2^2 = \langle  \left[\begin{smallmatrix} a & b \\ b& c\end{smallmatrix}\right],\bx\bx^{\top} \rangle$.
The key idea used in \cite{kileel2019expressivePolynomialNNs} is that for shallow networks ($L=2$) the order-$(K+1)$ tensor $\tensor{F}$ combining $\tensor{F}^{(i)}$ as $\tensor{F}_{i,:,\ldots,:} = \tensor{F}^{(i)}$ has a CPD with $d_1$ components (the numbers of hidden neurons):
\[
f(\bx)=\bA_2 \sigma_1(\bA_1 \bx)\,, 
\iff  \tensor{F} \emph{ has a CPD with factors } \bA_2, \underbrace{\bA_1^{\top},\ldots,\bA^{\top}_1}_{K \text{ times}}.
\]
This connection allows the use of results on generic properties of tensor decomposition to study both the expressivity and identifiability of PNNs.
In \cite{kileel2019expressivePolynomialNNs}, the authors show that for a sufficiently high number of hidden neurons ($d_1,\ldots,d_{L-1}$), the network is expressive, that is, it is able to represent any polynomial map (except for a set of Lebesgue measure zero).
This is done by lower bounding the dimension of the underlying algebraic variety to show that the variety is \emph{thick} or \emph{filling}. 
This dimension is also intimately linked to the presence or absence of bad local minima or spurious valleys in the optimization landscape when training under convex losses~\cite{kileel2019expressivePolynomialNNs}.

The parameter identifiability of such networks is a complementary question, as it can only hold when this dimension is \textit{sufficiently small} (similarly to tensors of subgeneric ranks).
It was shown in \cite{usevich2025identifiabilityDeepPNNs} that the identifiability of deep PNNs is intimately linked to the identifiability of shallow PNNs (and, thus, of tensor decompositions): a key result is that a deep ($L$-layer) polynomial NN is identifiable if every every 2-layer PNN subnetwork composed by a pair of two successive layers is also identifiable.

Other related work connecting the CPD to machine learning models exists.
Low-rank tensor formats have been used to reduce the number of parameters in the well-known \textit{factorization machines} \cite{rendle2010factorizationMachines}, which proposed to use the CPD to model a coefficient tensor describing the interaction between input features.
Symmetric tensors were also leveraged to unveil the identifiability properties of 2-layer polynomial NNs  with \emph{trainable activations} $f(\bx)=\bA_2\sigma_1(\bA_1\bx)$ (when  $\sigma_1(\cdot)$  is a vector of possibly different and arbitrary degree $K$ polynomials) through their connection to X-rank and coupled CP decompositions \cite{comon2017identifiabilityXrank}.

\subsection{\textbf{Sum-product networks: expressivity and approximation}}

\subsubsection{\textbf{Tensor decompositions as NNs}}
\begin{figure*}
\begin{mdframed}[backgroundcolor=black!7]
    \centering
    \begin{minipage}{0.39\linewidth}
    \fontsize{9pt}{10.5pt}\selectfont
    \textbf{Sum-product networks} (including polynomial NNs) are NN architectures directly connected to tensor decompositions, in which outputs are computed as the inner product between a \mbox{rank-1} tensor of feature representations of the input data $\bx$ and a coefficient tensor $\tensor{A}$, which can be parametrized in a low-rank format. This connection allows powerful \emph{expressivity}, \emph{approximation} and \emph{identifiability} results to be obtained.
    \end{minipage}%
    \hfill
    \begin{minipage}{0.5999\linewidth}
    \hfill
    \includegraphics[width=0.995\linewidth]{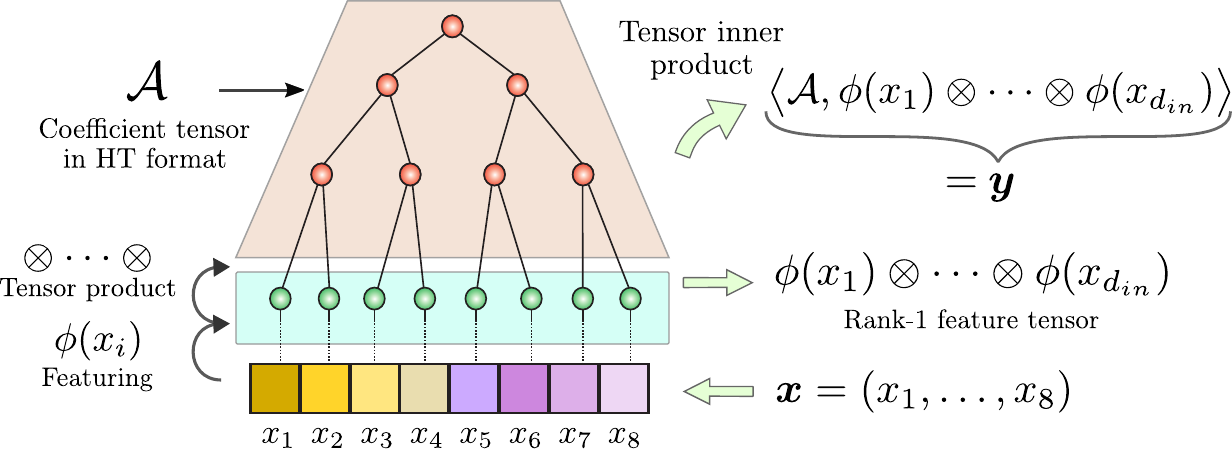} \hspace{-0.35cm}
    \end{minipage}
    \vspace{0.2cm}
    \caption{Illustration of the sum-product network~\eqref{eq:sum_product_net_tensor} with a coefficient tensor parametrized in HT~\cite{cichocki2017tensorNetworksPart1} format.}
    \label{fig:figure2_tensor_sumprodnets}
\end{mdframed}
\vspace{-1em}
\end{figure*}

In a more general context, hierarchical tensor decompositions are intimately linked to a class of neural network architectures called \emph{sum-product networks}, where the nonlinear activations $\sigma_{\ell}(\cdot)$ consist of products of their inputs \cite{cohen2016tensorExpressivePowerNeuralNets}. Their close connection to tensor decompositions unlocks a wealth of theoretical results to study their expressivity and approximation properties.
They key insight is that after some manipulations, these NN models (which include some forms of CNNs) can be expressed as (taking the scalar-valued case for simplicity):
\begin{align}
    \label{eq:sum_product_net_tensor}
    f(\bx) 
    &\,=\, \langle\tensor{A}, \phi(x_1) \otimes \cdots \otimes \phi(x_{\indim})\rangle
    \,=\, \sum_{k_1,\dots,k_{\indim}=1}^K \tensor{A}_{k_1,\dots,k_{\indim}} \prod_{i=1}^{\indim} \phi_{k_i}(x_{i}) \,,
\end{align}
where $\phi:\amsmathbb{R}\to\amsmathbb{R}^K$ is a feature extraction function which maps each input element to a feature space of dimension $K$ (examples include kernels, random feature expansion, or function tensorizations, which will be discussed below), with $\phi_{k}$ denoting its $k$-th output; $\tensor{A}\in\amsmathbb{R}^{K\times \dots\times K}$ is a cubic coefficient tensor of order $\indim$ and size $K$. 
This class of NNs is illustrated in Fig.~\ref{fig:figure2_tensor_sumprodnets}.

This model is very expressive, as it can capture interactions among any input features. However, the number of coefficients in $\tensor{A}$ grows exponentially with the input dimension $\indim$.
A key insight in \cite{cohen2016tensorExpressivePowerNeuralNets} was to link \eqref{eq:sum_product_net_tensor} to a type of convolutional sum-product NN architecture, and to study its expressivity. It was shown that while a CPD coefficient model is linked to a shallow NN, a special case of a HT decomposition with diagonal structure of inner factors  (e.g., diagonal $\tensor{G}^{(3)}$, $\tensor{G}^{(2)}_{i,:,:}$, $\tensor{G}^{(1)}_{i,:,:}$ in the example in \eqref{eq:HT})  corresponds to a deep network. Moreover, any generic function (i.e., with coefficients drawn from a continuous distribution) that can be implemented by a deep (HT) sum-product network of linear size would require exponential size to be realized by a shallow (CPD) network. In tensor language, this is related to the CP rank of a generic tensor being exponentially larger then its HT rank.

In a similar vein, a specific type of RNN with multilinear layers was shown to be equivalent to a sum-product model \eqref{eq:sum_product_net_tensor} in which the weight tensor $\tensor{A}$ following a TT decomposition \cite{khrulkov2018expressiveTensorNNs}. It was shown that for a generic RNN in TT format, an equivalent NN in CPD format would require exponentially larger width. This result is obtained by showing that a generic tensor in TT format with a given rank will have exponentially larger CP rank.

\subsubsection{\textbf{Studying expressivity of a single model for different ranks}}
A different type of result consists of looking at the expressivity of \eqref{eq:sum_product_net_tensor} not for different tensor formats (e.g., CP vs. TT), but for different rank values of the same decomposition. This reveals whether a higher ranks leads to more expressive models, and can show that there are values of rank beyond which no improvement in expressivity is obtained (which is generally the rank necessary to represent a generic tensor in the chosen format).

Sum-product network were shown to be equivalent to tensor networks, which are linked to many-body quantum systems \cite{levine2018deepQuantumEntanglementTensor}. 
This allows the use of notions from quantum mechanics to measure the expressivity of the such models, in particular the so-called \emph{quantum entanglement}, which is directly related to the amount of possible interactions between the input variables. The measure of quantum entanglement of a sum-product network is upper bounded by the ranks of the so-called \textit{matricizations} of the tensor network, which are essentially different rearrangements of the coefficient tensor as matrices.
Hence, there is a link between the expressivity of sum-product nets \eqref{eq:sum_product_net_tensor} and the generic rank of the associated tensor format. Thus, the effect of different choices of model architecture on the expressivity (such as the number of channels or the convolution stride) can be studied through their effect on the (upper bound of the) generic ranks of the tensor network~\cite{levine2018deepQuantumEntanglementTensor}.

In \cite{lizaire2024expressivityRNNsTensor}, the authors consider the effect of rank and latent dimension on the expressivity of second order RNNs, which are recurrent models that use a coefficient tensor in CP format to allow both first- and second-order interactions between hidden state and inputs at every iteration. In particular, varying the rank values establishes a strict hierarchy of expressivity in the resulting function class until a saturation 
point is reached (the point where the rank reaches the generic CP rank of the coefficient tensor), after which increasing the rank
 does not increase the expressivity of the second-order RNN model.

\begin{singlespace}
\begin{mytcolorbox}[title=\textbf{Expressivity results of sum-product networks from generic ranks of tensor formats}]

Once sum-product networks are represented in tensor format as in~\eqref{eq:sum_product_net_tensor}, the key notion to study their expressive power is through the generic propoerties of different low-rank formats (e.g., CPD, HT). These approaches can be divided in two categories:
\begin{itemize}
    \item \textbf{Comparing the ranks of two different tensor formats}: For a generic coefficient tensor admitting a CPD with rank $R$ (linked to a shallow network), what would be the rank that a HT or TT decomposition (linked to deep networks) would need to represent it? 
    Such results can help compare different NN architectures (e.g., shallow vs. deep)~\cite{cohen2016tensorExpressivePowerNeuralNets,khrulkov2018expressiveTensorNNs}.
    
    \item \textbf{Comparing a single tensor format for different ranks:} For a given format, does increasing the rank of the tensor increase flexibility/expressiveness? This is true for certain ranks, but not beyond a certain threshold: if the rank is sufficiently high (linked to non-unique decompositions), adding  components does not increase expressiveness \cite{lizaire2024expressivityRNNsTensor}.
\end{itemize}

\end{mytcolorbox}
\end{singlespace}

\subsubsection{\textbf{Tensorization and approximation theory in terms of smoothness spaces}}
While expressivity results are valuable to compare different architectures, they do not shed light into the smoothness properties of the associated function classes and their relation to classical approximation results.
Recently, significant effort was dedicated to understand the approximation spaces of NNs, such as ReLU nets, within the same setting as classical approximation results obtained in signal processing for polynomials, splines or wavelets, all of which are closely related to their smoothness properties \cite{gribonval2022approximationDeepNNs}.

In \cite{ali2023approximationTreeTensorNetworks} such analysis was performed by the tensorization of univariate functions $f:[0,1)\to\amsmathbb{R}$ over an interval, by the so-called \emph{coarse-graining}.
The idea is to use a partition of $[0,1)$ into $2^Q$ intervals of length $2^{-Q}$, and identify a one-dimensional function $f(x)$ with a multivariate function (or tensor) $f(\cdot,\ldots,\cdot,v)$,
so that $f(i_1,\ldots,i_Q,v)$, for $i_q \in \{0,1\}$ represent a restriction of the function on each sub-interval.
 Formally, by  using the dyadic expansion we can write $x=\sum_{q=1}^Q 2^{-k}i_q+2^{-Q}v$, for $i_q\in\{0,1\}$ and $v\in[0,1)$.
Finally, the function $f(\cdot,\ldots,\cdot,v)$, can be represented in a tensor format, e.g.,  
\begin{align}
    f(i_1,\ldots,i_Q,v) 
    &\,=\, \sum_{k=1}^K \zeta_1^{(k)}(i_1)\dots\zeta_Q^{(k)}(i_Q)\psi^{(k)}(v)
    \,=\, \langle \tensor{1}, \zeta_1(i_1) \otimes \dots \otimes \zeta_Q(i_Q) \otimes \psi(v) \rangle \,,
    \label{eq:approximation_tensorized1d}
\end{align}
for some feature dimension $K$ and functions $\zeta_i:\{0,1\}\to\amsmathbb{R}^{K}$ and $\psi:[0,1)\to\amsmathbb{R}^{K}$, which can be seen as a (higher dimensional) feature representation of the scalar input $x$. $\tensor{1}$ is the tensor of ones. Since the inputs of $\zeta_i$ are binary, these functions are essentially defined by their evaluations over $\{0,1\}$, that is, if we look at function $\zeta_1(i_1) \otimes \dots \otimes \zeta_Q(i_Q)$ for all binary inputs $i_k$, it can be equivalently represented as a tensor of real valued coefficients. Thus, these functions form a tensor network of order $Q$ (which can be seen as the depth of the NN) and dimension $K$, which was parametrized using a TT decomposition.

Such networks are highly expressive: It was shown in \cite{ali2023approximationTreeTensorNetworks} that any function belonging to a Besov space can be approximated with optimal rate 
by a network \eqref{eq:approximation_tensorized1d} in TT format. On the other hand, the approximation classes of this TT format are not embedded into any Besov space: an arbitrary function from this TT network has no Besov smoothness unless the depth of the network (i.e., $Q$) is constrained \cite{ali2023approximationTreeTensorNetworks}.
This result reconciles the approximation spaces of tensorized univariate NNs in TT format with similar approximation results that were recently obtained for deep ReLU networks \cite{gribonval2022approximationDeepNNs}.

\begin{singlespace}
\begin{mytcolorbox}[title=\textbf{Expressivity results must be interpreted with care}]
Despite the insight they shed on the interplay between width vs. depth in NNs, it is worth noting that the expressivity analysis of sum-product networks does not necessarily generalizes directly to other NN architectures. For example, there are (a subset of positive measure of) deep CNNs with ReLU activations realizable by small shallow architectures \cite{cohen2016cnns_generalizedTensor_RelU}.
\end{mytcolorbox}
\end{singlespace}

\section{Using derivatives and moment tensors  to learn neural networks}
\label{sec:part3_derivativesAndMoments}

\begin{singlespace}
\begin{mytcolorbox}[title=\textbf{Summary}: Using moments or derivatives of a NN to reveal its parameters as factors of low-rank tensors.]

\begin{itemize}
    \item[+] Leads to both parameter identifiability and generalization results and to polynomial time (tensor decomposition-based) learning algorithms that can be applied to widely used NN architectures (MLPs, CNNs).
    \item[-] Requires access to derivatives of the NN or to score function (information about the input data distribution).
    \item[-] Most methods address 2-layer NNs, adapting them to the case of 3- or more layers is difficult.
\end{itemize}

\tcbsubtitle{\textbf{Key results leveraged from low-rank tensor decompositions:}}
\begin{itemize}
    \item Uniqueness of tensor decompositions (CPD, Paratuck).
    \item Stability of the decomposition to perturbations/errors is essential in establishing generalization bounds.
    \item Computability results for the CPD and polynomial time algorithms for decomposing symmetric tensors.
    \item \textbf{Commonly used tensor formats}: (symmetric and non-symmetric) CPD, Paratuck.
\end{itemize}

\end{mytcolorbox}
\end{singlespace}

The core idea behind an important class of NN learning algorithms is to differentiate the function we want to estimate in order to relate its parameters to the factors of a (unique) tensor. This approach served as a pillar of recent breakthroughs in the theory of NN learning. 
While most modern NN training is based on gradient descent algorithms, training methods based on derivatives or moments provide a complementary approach with support for strong theoretical guarantees. Indeed, while complexity theory often shows overly pessimistic results when applied to NN learning, derivative/moment approaches have been used to prove that NN training can be done in polynomial time for some architectures (e.g., 2- or 3-layer MLPs). In addition, derivatives-based learning approaches also provide identifiability results for a large class of NNs.
In fact, such approaches have long history in signal processing, with connections to the method of moments in independent component analysis and latent variable models \cite{anandkumar2014tensor} and decoupling approaches in non-linear system identification \cite{dreesen2015decouplingPolinomials_1stOrder}.
Using derivative-based techniques to study deep NN architectures and the learning algorithms used in practice is an exciting open problem that might shed light on why practical NN training succeeds despite being NP-hard in the worst case.

\subsubsection{\textbf{Example for a 2-layer NN}}
To visualize the connection between NNs and a tensor, it is instructive to consider the example of a 2-layer NN with a scalar output, $f(\bx)=\ba_2 \sigma(\bA_1 \bx + \bb_1)$. Computing the third order derivative of $f(\bx)$ gives us
\begin{align}
    \nabla_{\bx}^{(3)} f(\bx) = \sum_{i=1}^{d_1} \gamma^{(i)}(\bx) \, \ba_1^{(i)} \otimes \ba_1^{(i)} \otimes \ba_1^{(i)} \,,
    \label{eq:derivative_2layer_NN}
\end{align}
where $\ba_1^{(i)}$ denotes the $i$-th row of $\bA_1$ ordered as a column vector, $\gamma^{(i)}(\bx)$ denotes the $i$-th element of $\gamma(\bx)=\diag(\ba_{2}) \sigma_i'''(\bA_1\bx+\bb_{1})$, and $\sigma'''$ is the third derivative of the (elementwise) activations $\sigma$.

\begin{figure*}
\begin{mdframed}[backgroundcolor=black!7]
    \centering
    \begin{minipage}{0.39\linewidth}
    \fontsize{9pt}{10.5pt}\selectfont
    By computing \textbf{derivatives} or carefully designed \textbf{moments} of a neural network, one can construct tensors that reveal parameters of the NN as the factors of a (unique) low-rank tensor format (e.g., the CPD or Paratuck). Then, tensor decomposition can be used to estimate the parameters of the NN and to support strong theoretical guarantees for both the model's \emph{generalization} and for the \emph{identification} of its parameters.    
    \end{minipage}%
    \hfill
    \begin{minipage}{0.5999\linewidth}
    \hfill
    \includegraphics[width=0.995\linewidth]{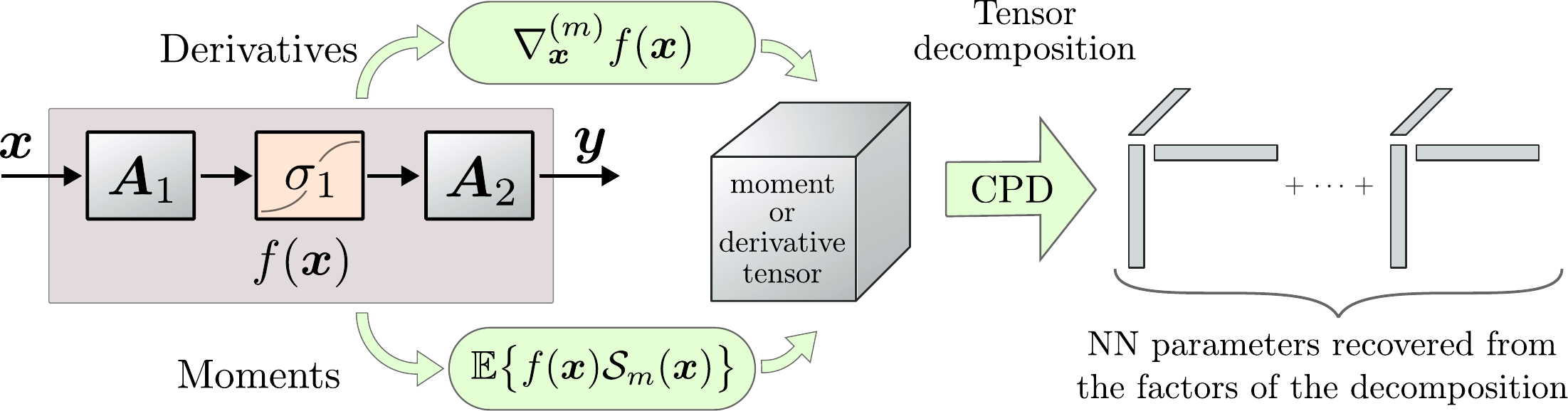} \hspace{-0.3cm}
    \end{minipage}
    \vspace{0.25cm}
    \caption{Illustration of differentiation-based methods (derivatives and method of moments) for learning a 2-layer NN.}
    \label{fig:figure3_tensor_derivatives}
\end{mdframed}
\vspace{-1em}
\end{figure*}

Inspecting this equation shows that the derivative \emph{reveals the parameters of the model in the form of a tensor}: the weight matrix $\bA_1$ is the factor of a symmetric CPD, whose rank equals the number of hidden neurons $d_1$. 
Thus, $\bA_1$ can be recovered by computing the symmetric CPD of $\nabla_x^{(3)} f(\bx)$. This allows us to leverage both decomposition algorithms as well as theoretical results (such as uniqueness), and already suggests the CPD as an initial step in an algorithm for NN learning. %
This is illustrated in Fig.~\ref{fig:figure3_tensor_derivatives}.

\subsubsection{\textbf{Score function approaches}}
One difficulty with exploiting \eqref{eq:derivative_2layer_NN} is the need to differentiate $f(\bx)$. Nonetheless, we can still construct tensors to train NNs using knowledge about the input data distribution $p(\bx)$ by means of the so-called \emph{score function} $\tensor{S}_m(\bx)=(-1)^m[\nabla^m p(\bx)]/p(\bx)$, which is an order-$m$ tensor of size $\indim\times ...\times \indim$ proportional the derivative of $\log p(\bx)$ \cite{anandkumar2014tensor,janzamin2015beatingPerilsTensorNeuralNets}. 
The key idea comes from a generalization of the classical Stein's lemma, %
which relates the (cross) moments between any continuously differentiable labeling function $f(\bx)$ and the the score function $\tensor{S}_m(\bx)$, to the expectation of the $m$-th order derivative of $f(\bx)$ as
\begin{align}
    \Ex_{\bx\sim p(\bx)}\big\{f(\bx) \otimes \tensor{S}_m(\bx)\big\} = \Ex_{\bx\sim p(\bx)}\big\{\nabla_{\bx}^{(m)} f(\bx)\big\} \,,
    \label{eq:steins_lemma}
\end{align}
under some mild regularity conditions. 
This result connects the derivative approach to the method of moments: the left hand side of \eqref{eq:steins_lemma} can be estimated from the data distribution, while and the right hand side give us the derivatives that can be used to connect the NN weights to a tensor decomposition. Thus, \emph{we forego the need to evaluate the derivatives of $f(\bx)$ by assuming access to the score function if we consider its average behavior}. This served as a basis of theoretical results and NN learning algorithms~\cite{janzamin2015beatingPerilsTensorNeuralNets}.

\begin{singlespace}
\begin{mytcolorbox}[title=\textbf{From derivatives to score functions and the method of moments}]

The derivatives are an essential step to reveal the parameters of a NN through tensor decompositions. When leveraging Stein's lemma \eqref{eq:steins_lemma}, we exchange the need for knowing how to differentiate $f(\bx)$ by the knowledge of the score function $\tensor{S}_m(\bx)$. This shows that knowledge of the input data distribution can make NN learning more tractable. However, \emph{the score function can be very hard to estimate without further assumptions} on $p(\bx)$. 
When the input is a standard Gaussian $p(\bx)=\gaussian(\cb{0},\bI)$, then the score functions are given by Hermite polynomials $\mathcal{S}_m(\bx)=\hermite_k(\bx)$. This result has been used to develop efficient NN learning algorithms based on the method of moments.

\end{mytcolorbox}
\end{singlespace}

\begin{singlespace}
\begin{mytcolorbox}[title=\textbf{Learning of 2-layer NNs $f(\bx)=\bA_2 \sigma(\bA_1 \bx + \bb_1)$ using the method of moments or derivatives}]%

\begin{enumerate}
    \item Estimate cross-moments or derivative tensors, as $\Ex_{\bx\sim p(\bx)}\big\{f(\bx) \otimes \tensor{S}_3(\bx)\big\}$ or $\Ex_{\bx\sim p(\bx)}\big\{\nabla_{\bx}^{(3)} f(\bx)\big\}$.
    
    \item Identify the factors of the tensor with some NN parameters (e.g., the first layer weights $\bA_1$).
    
    \item Compute these parameters using tensor decomposition (e.g., symmetric CPD).
    
    \item Given the computed factors (e.g., $\bA_1)$, recover the remaining NN parameters (e.g., bias $\bb_1$, activation functions $\sigma(\cdot)$) using, e.g., Fourier-based and/or regression techniques.
\end{enumerate}

\end{mytcolorbox}
\end{singlespace}

\subsection{\textbf{Learning 2-layer NNs with the method of moments}}

The method of moments was first leveraged in \cite{janzamin2015beatingPerilsTensorNeuralNets} for learning 2-layer NNs, leading to a polynomial-time algorithm and an in-depth theoretical analysis.
Consider a network $f(\bx)=\ba_2 \sigma(\bA_1 \bx + \bb_1)$ with bias and sigmoidal activation $\sigma$ with nonzero (expected) third derivatives. Leveraging the connection between the order-3 score function and the NN derivatives, a general algorithm was devised in four key steps:
\begin{enumerate}
    \item Estimate the cross-moments between the NN output and the input score function in \eqref{eq:steins_lemma}.
    
    \item Estimate the weights of the first layer $\bA_1$ using the symmetric CP decomposition of $\Ex_{\bx\sim p(\bx)}\big\{\nabla_{\bx}^{(3)} f(\bx)\big\}$.
    
    \item Given the estimated $\bA_1$ and knowledge of $p(\bx)$, estimate the bias term $\bb_1$ using a Fourier-transform method. Specifically, since the transformation \mbox{$\ba_2 \sigma(\bA_1 \bx)\mapsto \ba_2 \sigma(\bA_1 \bx + \bb_1)$} is a shift of the function $\ba_2 \sigma(\bA_1 \bx)$, $\bb_1$ can be recovered  uniquely from the phase of its Fourier transform. %
    
    \item Finally, the parameters of the last layer $\ba_2$ can be estimated by least-squares regression.
\end{enumerate}

Assuming knowledge of $\mathcal{S}_3(\bx)$ and some non-degeneracy conditions on $\bA_1$, this algorithm achieves a polynomial sample complexity (scaling as $1/\sqrt{N}$ with the number of training samples $N$) both for the model's generalization and for the recovery of its parameters, which are uniquely identified \cite{janzamin2015beatingPerilsTensorNeuralNets}. 
The analysis combines statistical estimation bounds to account for estimation errors in the moments due to a finite number of samples with a perturbation analysis of a symmetric CPD algorithm.

Several subsequent works investigated the method of moments to learn 2-layer NNs by exploiting more properties of the data and NN architecture. 
The authors in \cite{ge2019learning2layerNNsSymmetricInputs} considered 2-layer NNs without bias, with symmetric input distribution $p(\bx)=p(-\bx)$, ReLU activations $\sigma(x)=\max(0,x)$ and with the output dimension being the same as the number of hidden neurons (i.e., $\bA_2$ being square). In this case, the score function is not directly necessary: well-chosen (cross) moments (of order up to four) between inputs $\bx$ and outputs $\by$ can be directly related to the the NN parameters, which can be estimated using tensor decomposition. 
This approach was extended in \cite{awasthi2021efficient2LayerNNRelu} to address ReLU NNs with bias and scalar outputs by assuming the input distribution $p(\bx)$ to be Gaussian using the tensor decomposition of the so-called Hermite coefficients of the NN $f(\bx)$ (i.e., the score function for the Gaussian case, $\Ex\{f(\bx) \hermite_k(\bx)\}$). %

\subsubsection{\textbf{Convolutional architectures}}
The method of moments has also been used in to learn the parameters of convolutional network architectures in~\cite{oymak2021learningCNNsTensorsMoM}, where a particular tensorization of the data was exploited to construct a tensor with the convolution kernels as its factors from the first order score function. However, the convolution kernels are learned by non-symmetric CP decomposition, which is generally NP-hard.

\subsubsection{\textbf{Connecting the method of moments to optimization approaches}}
Tensorial approaches have been used in tandem with optimization strategies to train NNs. By analyzing the least squares loss function for learning 2-layer NNs with Gaussian inputs and no bias terms, it was shown in \cite{ge2018learning2layerNNsLandscapeDesign} that the optimization loss is implicitly equivalent to simultaneously decomposing a sequence of low-rank tensors, weighted by the Hermite coefficients of the activation function $\xi_k$, that is, $\Ex_{\bx\sim p(\bx)}\{|f(\bx)-\hat{f}(\bx)|^2\}=\sum_{k=1}^{\infty} \xi_k\|\sum_{r=1}^{d_1} \ba_r^{\otimes k} - \sum_{r=1}^{d_1}\hat{\ba}_r^{\otimes k}\|^2$ for two networks $f$ and $\hat{f}$. Using this connection, the authors modified the loss function (effectively changing the activation function) to obtain more favorable optimization landscapes. 
The method of moments can be also used to provide sufficiently accurate initializations to stochastic gradient algorithms such that convergence to a global optimum could be guaranteed.%

\begin{singlespace}
\begin{mytcolorbox}[title=\textbf{Score matching/method of moments:}]
\begin{itemize}
    \item[+] No need to probe the (unknown) function $f(\bx)$ to compute derivatives.
    \item[-] need access to the score function $\tensor{S}_m(\bx)$ (information about the input data distribution).
    \item[+] Often leads to symmetric tensors that can be tackled with well-mastered (polynomial time) algorithms.
    \item[+] Very efficient for some choices of input distribution (Gaussian, symmetric) and activations (ReLU).
\end{itemize}
\tcbsubtitle{\textbf{Direct differentiation:}}
\begin{itemize}
    \item[+] Can tackle trainable activation functions and 3-layer NNs.
    \item[-] Leads to non-symmetric or non-additive tensor decompositions (NP-hard).
    \item[-] Needs access to the derivatives of function $f(\bx)$ to be learned (e.g., in the active sampling regime).
\end{itemize}

\end{mytcolorbox}
\end{singlespace}

\subsection{\textbf{Using direct differentiation and flexible activation functions}}

Several approaches are based on directly evaluating derivative matrices and tensors (Jacobians and Hessians) and using their connection to the NN architecture for learning both weight matrices and trainable activation functions \cite{dreesen2015decouplingPolinomials_1stOrder,fornasier2021robustIdentificationShallowNeuralNets}. 
Supposing the output of a (vector-valued) NN $f(\bx)=\bA_2\sigma(\bA_1\bx+\bb_1)$ can be evaluated at points selected by the user (e.g., experimental design or compression of preexisting NNs), a third-order tensor $\tensor{J}\in\amsmathbb{R}^{\outdim\times \indim \times N}$ can be constructed by stacking evaluations of the Jacobian matrix of $f(\bx)$ at $N$ different points $\bx_1,\ldots,\bx_N$ as \cite{dreesen2015decouplingPolinomials_1stOrder}
\[
    \tensor{J}_{:,:,n} \,\triangleq\, \nabla_{\bx}^{(1)} f(\bx_n) \,=\, \bA_2 \diag\big(\sigma'(\bA_1\bx_n+\bb_1)\big) \bA_1 \,,
\]
where $\sigma'$ contains the first-order derivatives of the activation functions in $\sigma$. This tensor admits a CP decomposition with rank equal to the number of neurons. Its factors in modes 1 and 2 are the NN weight matrices $\bA_1$ and $\bA_2$, and the mode-3 factor consists of different evaluations of $\sigma'$, which can be recovered uniquely under mild assumptions~\cite{dreesen2015decouplingPolinomials_1stOrder}.
Differently from most approaches based on the method of moments, decoupling methods use first-order information. However, this leads to the CPD of non-symmetric tensors, which is harder to compute (and NP-hard in general).

This does not assume knowledge of $\sigma$, thus, the (possibly different) activation functions (and bias term) can be learned by choosing a parametric form for $\sigma\triangleq\sigma_{\theta}$ depending on parameters $\theta$, and fitting its derivative $\sigma_{\theta}'$ to the the recovered mode-3 factors $\bc^{(n)}\triangleq\sigma'(\bA_1\bx_n+\bb_1)$, using $\bA_1\bx_n$ as inputs~\cite{dreesen2015decouplingPolinomials_1stOrder}, which makes the model very flexible.
Common choices for $\sigma_{\theta}$ include polynomials or expansions in some basis of functions chosen a priori.

\begin{singlespace}
\begin{mytcolorbox}[title=\textbf{When can NNs be learned in polynomial time? Is the answer related to tensor decompositions?}]

Algorithms based on the method of moments have provided encouraging answers~\cite{janzamin2015beatingPerilsTensorNeuralNets,ge2018learning2layerNNsLandscapeDesign}, relying on the tractability of symmetric CPD. However, many approaches (dealing with trainable activations~\cite{dreesen2015decouplingPolinomials_1stOrder} or CNNs~\cite{oymak2021learningCNNsTensorsMoM}) use non-symmetric CPD, which is generally NP-hard. 
Even in the symmetric case the situation is intricate. Let us take as example 2-layer NNs $f(\bx)=\cb{1}^\top\sigma(\bA_1\bx)$ in which $\sigma(\cdot)$ are third degree polynomials, $\bx\sim\gaussian(\cb{0},\bI)$ and $\bA_1$ is drawn from some isotropic distribution. It was shown in \cite{mondelli2019connection2layerNnetTensorDec} that the hardness of learning this NN depends on the (in)existence of efficient algorithms to decompose tensor $\sum_{i=1}^{d_1} \ba_1^{(i)} \otimes \ba_1^{(i)} \otimes \ba_1^{(i)}$, which is conjectured to be NP-hard on $\indim$ when the number of neurons/rank satisfies $d_1\geq (\indim)^{3/2}$. Moreover, under some further assumptions, when the number of neurons satisfies $(\indim)^{3/2}<d_1<(\indim)^2$, the expected risk for $f$ given by any polynomial time algorithm is bounded away from zero with high probability when $\indim\to\infty$.
This suggests a possible phase transition of learnability at $(\indim)^{3/2}$ neurons, and illustrates a deep connection between tensors and NNs.

\end{mytcolorbox}
\end{singlespace}

\subsubsection{\textbf{Connection to active subspace approaches}}
Approaches based on derivatives are connected to \emph{active subspace} (AS) methods, which are nonlinear dimension reduction approaches that identify a set of important directions in the input space of a scalar-valued $f(\bx)$ by projecting $\bx$ onto the column space of the averaged gradient matrix $\Ex_{\bx\sim p(\bx)}\{(\nabla_{\bx}f(\bx))(\nabla_{\bx}f(\bx))^\top\}$.
AS approaches have been used to learn NNs with less neurons than input dimensions ($d_1\leq \indim$) with robustness to perturbations in the derivative estimation~\cite{fornasier2021robustIdentificationShallowNeuralNets}. Considering a NN $f(\bx)=\cb{1}^\top \sigma(\bA_1\bx))$, the authors proposed the following algorithm:
\begin{enumerate}
    \item Use the AS approach to reduce the dimension to the input space to $d_1$.
    \item Use averaged gradient and Hessian information of $f(\bx)$ to estimate $\vspan\{\ba_1^{(i)}\}_i\subset\amsmathbb{R}^{d_1}$ and $\vspan\{\ba_1^{(i)}\otimes\ba_1^{(i)}\}_i\subset\amsmathbb{R}^{d_1\times d_1}$, and decompose these subspaces jointly using a non-convex optimization problem to recover $\bA_1$ (up to permutations and sign changes).
    \item Once $\bA_1$ is estimated, the activations functions $\sigma$ can be learned using regression.
\end{enumerate}
Theoretical guarantees are provided for the correct recovery of $\bA_1$ and for the generalization of $f(\bx)$, accounting for errors in derivative estimation~\cite{fornasier2021robustIdentificationShallowNeuralNets}. The joint use of first- and second-order differentials was shown to improves the sensitivity to noise compared to using only third-order derivatives.

\begin{singlespace}
\begin{mytcolorbox}[title=\textbf{A framework to study model identifiability} ]

A key feature of algorithms based on derivatives or on the method of moments is that they inherit uniqueness properties of tensor decompositions such as the CPD. This can lead to results guaranteeing that the network not only generalizes well (i.e., the expected risk is small), but also that the NN parameters/weights can be recovered uniquely (up to trivial ambiguities), ensuring the interpretation (e.g., disentanglement) of the hidden representations. 

\end{mytcolorbox}
\end{singlespace}

\subsubsection{\textbf{Learning 3-layer NNs}}
The majority of results for learning NNs using derivatives address the case of 2-layers. Nonetheless, both derivative- and AS-based approaches have been extended to the 3-layer case \cite{dejonghe2023twolayer,fornasier2022robustIdentification2layerNeuralNets}. %
For a NN $f(\bx)=\bA_3\sigma_2(\bA_2\sigma_1(\bA_1\bx))$ with $\outdim>1$, one approach is to, again, take derivatives to construct a Jacobian tensor which now will have the following format~\cite{dejonghe2023twolayer}:
\begin{align}
    \nabla_{\bx}^{(1)} f(\bx_n) = \bA_3 \diag\big(\sigma_2'(\bz_n)\big) \bA_2 \diag\big(\sigma_1'(\bt_n)\big) \bA_1 \,,
\end{align}
where $\bt_n=\bA_1\bx_n$ and $\bz_n=\bA_2\sigma_1(\bA_1\bx_n)$. This is no longer an additive decomposition, but a Paratuck-2 decomposition \cite{kolda2009tensor}. This result was used in \cite{dejonghe2023twolayer} to devise a learning algorithm based on the factorization of the Jacobian of $f(\bx)$.

A different approach based on AS was proposed in \cite{fornasier2022robustIdentification2layerNeuralNets} to learn 3-layer NNs of the form $f(\bx)=\cb{1}^\top\sigma_2(\bA_2\sigma_1(\bA_1\bx))$. Inspecting the Hessian of $f(\bx)$ immediately reveals the key challenge:
\begin{align}
    \nabla_{\bx}^{(2)} f(\bx) = \sum_{i=1}^{d_1} \tilde{\gamma}^{(i)}(\bx) \, \ba_1^{(i)} \otimes \ba_1^{(i)} 
    + \sum_{j=1}^{d_2} \eta^{(j)}(\bx) \, \bv^{(j)}(\bx) \otimes \bv^{(j)} (\bx) \,,
\end{align}
where $\tilde{\gamma}^{(i)}(\bx)$ and $\eta^{(j)}(\bx)$ are coefficient functions, and $\bv^{(j)}(\bx)$ is a factor of the decomposition which is \emph{no longer constant, but a function of the inputs $\bx$}. For suitable statistical distributions of sampling points $\bx_n$, the span of the Hessians $\big\{\nabla_{\bx}^{(2)} f(\bx_n)\big\}_{n=1}^N$ concentrate around the fixed subspace
$\vspan\big\{{\ba_1^{(i)}\otimes \ba_1^{(i)}}, \, \bv^{(j)}(\cb{0}) \otimes \bv^{(j)}(\cb{0}) \big\}_{i=1,\ldots,d_1, j=1,\ldots,d_2}$,
where $\bv^{(j)}(\cb{0})$ is a known function of the NN parameters. This relation allows the recovery of the NN weights by solving a nonlinear optimization problem \cite{fornasier2022robustIdentification2layerNeuralNets}.

\begin{singlespace}
\begin{mytcolorbox}%
\textbf{Going beyond 2-layer NNs} with derivative or moment-based methods is a challenging endeavor as we lose the link to standard tensor decompositions: the gradients of are no longer related to a CPD but to non-additive (e.g., Paratuck) decompositions, and the Hessian has a nonlinear structure.
\end{mytcolorbox}
\end{singlespace}

\section{Emerging use of tensors in other learning problems}
\label{sec:part4_TensorsInotherLearningProblems}

\begin{singlespace}
\begin{mytcolorbox}[title=\textbf{Summary:} {Use of tensors to advance the theory of RL, generative modeling, and mixtures of classifiers.}]
\begin{itemize}
    \item Link between generative models with polynomial NNs and tensor ring decomposition, with polynomial time learning algorithms under some dimension settings based on the method of moments.
    
    \item Action-value functions in RL with discrete state and action spaces can be represented as a tensor with a low-rank format. 
    
    \item Mixtures of linear classifiers can be identified using CPD based on moments of a carefully chosen distribution.
\end{itemize}
\tcbsubtitle{\textbf{Key results leveraged from low-rank tensor decompositions:}}
\begin{itemize}
    \item Uniqueness of the CPD, compression of the amount of parameters, efficient algorithms to learn tensor networks.
    \item \textbf{Commonly used tensor formats}: TT/tensor ring, CPD.
\end{itemize}
\end{mytcolorbox}

\end{singlespace}

Recently, the tensorial approaches described in the previous sections have been extended to study more general learning problems in generative modeling, RL, and learning mixtures of linear classifiers.

\subsubsection{\textbf{Generative models}}
Tensors have been used to study the learnability of generative models in~\cite{chen2023learningPolynomialGenerativeModelsTensors}. For Gaussian inputs $\bx\sim\gaussian(\cb{0},\bI)$, one can consider the pushforward measure $\by=f(\bx)$ on $\amsmathbb{R}^{\outdim}$ when $f$ is a polynomial NN of degree $K$, whose $i$-th output can be written as
\[y^{(i)} = \langle \tensor{A}^{(i)}, \bx^{\otimes K}\rangle \,, \quad i=1,\ldots,\outdim\,,\]
where $\tensor{A}^{(i)}\in\amsmathbb{R}^{\indim\times\ldots\times\indim}$ are parameter tensors. The particularity is that $\{\tensor{A}^{(i)}\}_{i=1}^{\outdim}$ are learned only from data $\{\by_n\}_{n=1}^N$, without access to the inputs $\{\bx_{n}\}_{n=1}^N$. 
An algorithm based on the method of moments (using moments of order $\le3$) was proposed in~\cite{chen2023learningPolynomialGenerativeModelsTensors} and shown to be equivalent to a tensor ring decomposition \cite{cichocki2017tensorNetworksPart1}. Moreover, quadratic NNs ($K=2$) can be learned in polynomial time when $\indim = O(\sqrt{\outdim})$.

Another important use of tensors is in learning discrete hidden Markov models (HMM). An HMM with discrete hidden variables $h_i\in\{1,\ldots,H\}$ can be written as
\[p(x_1,\ldots,x_{\indim}) = \sum_{h_1,\ldots,h_{\indim}=1}^H p(x_1|h_1)\prod_{i=1}^{\indim} p(h_i|h_{i-1})p(x_i|h_i) \,,\]
which is equivalent to a nonnegative tensor network model \cite{glasser2019expressiveTensorNetworksProbabilisticModeling}. This connection has been used to study the expressivity of such models following a similar strategy as explained in Section~\ref{sec:part2_NNsAsTensorsSumProduct}. Similar connections can also be established between tensor networks and restricted Boltzmann machines. %

\subsubsection{\textbf{Mixtures of linear classifiers}}
An emerging problem in machine learning is learning a mixture of $M$ linear classifiers \cite{chen2022mixtureLinearClassifiersTensor}.
Given a set of vectors $\ba_1,\ldots,\ba_M\in\amsmathbb{R}^{\indim}$ and corresponding probability weights $w_j$, in this setting the measured data is generated as follows: an index $j$ is first selected with probability $w_j$, and then the labels are computed as $y=\iota_+(\ba_j^\top\bx)$, where $\bx\sim\gaussian(\cb{0},\bI)$ and $\iota_+$ is the indicator function of $\amsmathbb{R}_+$. We aim to recover $\ba_j$ and $w_j$. 
Tensor decomposition has been used in \cite{chen2022mixtureLinearClassifiersTensor} to provide a learning algorithm based on the method of moments that guarantees identifiability of the parameters with with polynomial dependence on $\indim$ and exponential dependence on $M$ and on the separation between weight vectors, $\|\ba_i-\ba_j\|$, $i\neq j$. The key insight is to circumvent the nonlinearity in $\iota_+$ by estimating the moments not of $p(\bx)$, but of $p(\bx|y\geq 0)$. This reveals the parameters of the model as a tensor, since
\[\Ex_{\bx\sim p(\bx|y\geq 0)}\{\hermite_{2D+1}(\bx)\} \mypropto \sum_{j=1}^M w_j \ba_j^{\otimes (2D+1)} \,.\]
The parameters can then be uniquely estimated using CP tensor decomposition for high enough~$D$.

\subsubsection{\textbf{Reinforcement learning}}
In a nutshell, in RL, an agent interacts with an environment over time $t=1,2,\ldots$ by selecting actions $\bu_t\in \mathscr{U}$ depending on its current state $\bs_t\in\mathscr{S}$ (which transitions according to the distribution $p(\bs_{t+1}|\bs_t,\bu_t)$) in order to maximize a (expected and accumulated) reward, measured through the action-value function $Q(\bs_t,\bu_t) = \Ex\{\sum_{k=0}^{\infty}\gamma^{k}r(\bs_{t+k},\bu_{t+k})|\bs_t,\bu_t\}$, for some discount factor $\gamma\in[0,1)$. Given a policy $\pi(\bu|\bs)$, which specifies the probability of selecting actions given a state, a cornerstone of RL frameworks is the Bellman expectation equation
\[Q(\bs,\bu) = r(\bs,\bu) + \gamma\Ex_{p(\bs'|\bs,\bu)\pi(\bu'|\bs') }\{Q(\bs',\bu')\} \,,\]
which characterizes the optimal action-value function $Q(\bs,\bu)$ that needs to be recursively estimated. A key observation is that when the action and state spaces $\mathscr{U}$ and $\mathscr{S}$ are discrete and have high dimension, $Q(\bs,\bu)$ can be represented as a tensor consisting of the evaluations of this function over the possible values of $\bs$ and $\bu$. This prompted the use of low-rank decomposition to parametrize $Q(\bs,\bu)$, along with recursive strategies for its online estimation \cite{mahajan2021tesseract}. It was theoretically shown that the rank of the tensorized $Q(\bs,\bu)$ can be bounded in terms of the ranks of the reward and transition tensors, and that the use of low-rank assumptions considerably reduces the sample complexity \cite{mahajan2021tesseract}.

\section*{Conclusions and perspectives}

This article provided an overview of the different ways in which the framework of low-rank tensor decompositions has been exploited in the theoretical study of NNs. The intimate connection between tensor formats and NNs supported the study of the compression of NN architectures, of tensorized (sum-product) networks, and of learning using derivatives and moments, addressing questions such as the expressivity, generalization, learnability, and identifiability of NNs.
As the use of deep learning advances at an impressive speed, important challenges continue to emerge, among which:

\textbf{Understanding NNs with low-rank weights:} Few works managed to investigate the impact of low-rank weights on the generalization of NNs. Understanding the influence of different low-rank formats (including implicit biases of NN training algorithms towards solutions with low rank) and their use in emerging NN architectures such as transformers and state-space models is of fundamental importance.

\textbf{Approximation and tensorized neural networks:} The connection between sum-product networks and tensorized univariate functions and tensor decompositions such as TT and HT allowed powerful expressivity and approximation results to be obtained. %
How much can these results tell us about other classes of architectures, such as deep transformers? Tightening the connection between these different families of neural networks can further the understanding of a broader class of deep learning models.

\textbf{Derivatives-based methods beyond the 2-layer or Gaussian cases:} Methods based on derivatives and the method of moments have supported approaches to learn NNs with strong
guarantees. However, most algorithms are restricted to the 2-layer case or need knowledge about the distribution of the input data. Extending these methods to efficiently handle deep networks and unknown input data distributions is essential to broaden the use of these techniques.

The interface between the mathematics of low-rank tensor decompositions and the theory of NNs proved to be a fruitful area of research. We hope that the interaction between these fields will continue to boost important contributions, both in the theoretical study of well-established and emerging tensor formats and in their use to investigate cutting edge NN architectures.

\section*{Acknowledgment}
This work was supported in part by the French National Research Agency (ANR) under grants ANR- 23-CE23-0024, ANR-23-CE94-0001, by the PEPR project CAUSALI-T-AI, and by the National Science Foundation, under grant NSF 2316420.

\begin{singlespace}
\bibliographystyle{IEEEtran}
\bibliography{references}
\end{singlespace}

\end{document}

%% file: MathSymbolDefs2.tex
\usepackage{amsmath,amssymb,mathrsfs,bbm}
\usepackage{graphicx}
\usepackage[colorinlistoftodos]{todonotes}
\usepackage{multirow}
\usepackage{makecell}

\usepackage{psfrag,epsfig,graphics}
\usepackage{amsmath,amsthm,amssymb,multirow}
\usepackage{mathbbol}
\usepackage{amssymb}   
\usepackage{url}  
\usepackage{xstring}

\usepackage{mathabx} %

\DeclareSymbolFontAlphabet{\amsmathbb}{AMSb}%

\newcommand{\cp}[1]{\ifmmode {\mathcal{#1}}\else ${\mathcal{#1}}$\fi}
	
\newcommand{\bA}{\boldsymbol{A}}
\newcommand{\bB}{\boldsymbol{B}}
\newcommand{\bC}{\boldsymbol{C}}

\newcommand{\bI}{\boldsymbol{I}}

\newcommand{\bU}{\boldsymbol{U}}
\newcommand{\bV}{\boldsymbol{V}}
\newcommand{\bW}{\boldsymbol{W}}
\newcommand{\bX}{\boldsymbol{X}}

\newcommand{\ba}{\boldsymbol{a}}
\newcommand{\bb}{\boldsymbol{b}}
\newcommand{\bc}{\boldsymbol{c}}

\newcommand{\bh}{\boldsymbol{h}}

\newcommand{\by}{\boldsymbol{y}}
\newcommand{\bs}{\boldsymbol{s}}
\newcommand{\bt}{\boldsymbol{t}}
\newcommand{\bu}{\boldsymbol{u}}
\newcommand{\bv}{\boldsymbol{v}}
\newcommand{\bx}{\boldsymbol{x}}

\newcommand{\bz}{\boldsymbol{z}}

\newcommand{\bbR}{\amsmathbb{R}}

\newcommand{\cb}[1]{\boldsymbol{#1}}

\newcommand\tensor[1]{%
  \ifcat\noexpand#1\relax %
    \mathbb{#1}%
  \else
      \if\relax\detokenize\expandafter{\romannumeral-0#1}\relax  %
        \mathbb{#1}
      \else
        \mathcal{#1}%
      \fi
  \fi }

\newcommand{\diag}{\operatorname{diag}}

\newcommand{\vspan}{\operatorname{span}}

\newcommand{\Ex}{\amsmathbb{E}}

\newcommand{\tensorize}{\operatorname{Ten}}

\newcommand{\riskloss}{\mathsf{L}}

\newcommand{\hermite}{\mathsf{He}}

\newcommand{\genrank}{r_{\rm gen}}
\newcommand{\maxrank}{r_{\rm max}}

\newcommand{\indim}{{d_{\rm in}}}
\newcommand{\outdim}{{d_{\rm out}}}
\newcommand{\gaussian}{\mathscr{N}}

\newcommand{\mypropto}{\,\propto\,}

\usepackage{utfsym}
\newcommand{\newcheckmark}{\usym{2713}}
\newcommand{\newcrossmark}{\usym{2717}}

\usepackage{color}  %
\usepackage{colortbl}  %

\definecolor{darkgreen}{rgb}{0., 0.4, 0.}

\usepackage[most]{tcolorbox} 

\newcommand{\fontsizeninepointfive}{\fontsize{9.5pt}{11pt}\selectfont}

\newtcolorbox[auto counter, number within=section]{mytcolorbox}[1][]{ 
  add to width=1cm,               %
  enlarge left by=0.5cm,          %
  enlarge right by=0.5cm,         %
  center,                         %
  left=2pt,                       %
  colframe=orange!60!white,   %
  colback=gray!18!white,          %
  colbacktitle=orange!60!white,   %
  coltitle=black,                 %
    fontupper=\fontsizeninepointfive,
        fonttitle=\small,  %
  subtitle style={
    colback=orange!60!white,      %
    colupper=black,               %
    boxrule=0.5pt,                 %
  },
  #1                             %
}

\usepackage{mdframed}

%% file: SPM_revpaper_main_final.bbl
\begin{thebibliography}{10}
\providecommand{\url}[1]{#1}
\csname url@samestyle\endcsname
\providecommand{\newblock}{\relax}
\providecommand{\bibinfo}[2]{#2}
\providecommand{\BIBentrySTDinterwordspacing}{\spaceskip=0pt\relax}
\providecommand{\BIBentryALTinterwordstretchfactor}{4}
\providecommand{\BIBentryALTinterwordspacing}{\spaceskip=\fontdimen2\font plus
\BIBentryALTinterwordstretchfactor\fontdimen3\font minus
  \fontdimen4\font\relax}
\providecommand{\BIBforeignlanguage}[2]{{%
\expandafter\ifx\csname l@#1\endcsname\relax
\typeout{** WARNING: IEEEtran.bst: No hyphenation pattern has been}%
\typeout{** loaded for the language `#1'. Using the pattern for}%
\typeout{** the default language instead.}%
\else
\language=\csname l@#1\endcsname
\fi
#2}}
\providecommand{\BIBdecl}{\relax}
\BIBdecl

\bibitem{sidiropoulos2017tensor}
N.~D. Sidiropoulos, L.~De~Lathauwer, X.~Fu, K.~Huang, E.~E. Papalexakis, and
  C.~Faloutsos, ``Tensor decomposition for signal processing and machine
  learning,'' \emph{IEEE Transactions on signal processing}, vol.~65, no.~13,
  pp. 3551--3582, 2017.

\bibitem{comon2014spm}
P.~Comon, ``Tensors : A brief introduction,'' \emph{IEEE Signal Processing
  Magazine}, vol.~31, no.~3, pp. 44--53, May 2014.

\bibitem{kolda2009tensor}
T.~G. Kolda and B.~W. Bader, ``Tensor decompositions and applications,''
  \emph{SIAM review}, vol.~51, no.~3, pp. 455--500, 2009.

\bibitem{hackbusch2012tensorBook}
W.~Hackbusch, \emph{Tensor spaces and numerical tensor calculus}.\hskip 1em
  plus 0.5em minus 0.4em\relax Springer, 2012, vol.~42.

\bibitem{panagakis2021tensorsComputerVisionDeepLearning}
Y.~Panagakis, J.~Kossaifi, G.~G. Chrysos, J.~Oldfield, M.~A. Nicolaou,
  A.~Anandkumar, and S.~Zafeiriou, ``Tensor methods in computer vision and deep
  learning,'' \emph{Proc. IEEE}, vol. 109, no.~5, pp. 863--890, 2021.

\bibitem{anandkumar2014tensor}
A.~Anandkumar, R.~Ge, D.~Hsu, S.~M. Kakade, and M.~Telgarsky, ``Tensor
  decompositions for learning latent variable models,'' \emph{Journal of
  machine learning research}, vol.~15, pp. 2773--2832, 2014.

\bibitem{cichocki2017tensorNetworksPart1}
A.~Cichocki, N.~Lee, I.~Oseledets, A.-H. Phan, Q.~Zhao, D.~P. Mandic
  \emph{et~al.}, ``Tensor networks for dimensionality reduction and large-scale
  optimization: Part 1 low-rank tensor decompositions,'' \emph{Foundations and
  Trends{\textregistered} in Machine Learning}, vol.~9, no. 4-5, pp. 249--429,
  2016.

\bibitem{novikov2015tensorizingNNs}
A.~Novikov, D.~Podoprikhin, A.~Osokin, and D.~P. Vetrov, ``Tensorizing neural
  networks,'' \emph{Adv. Neur. Inf. Proc. Syst.}, vol.~28, 2015.

\bibitem{hu2022lora}
E.~J. Hu, Y.~Shen, P.~Wallis, Z.~Allen-Zhu, Y.~Li, S.~Wang, L.~Wang, and
  W.~Chen, ``Lo{RA}: Low-rank adaptation of large language models,'' in
  \emph{International Conference on Learning Representations (ICLR)}, 2022.

\bibitem{cohen2016tensorExpressivePowerNeuralNets}
N.~Cohen, O.~Sharir, and A.~Shashua, ``On the expressive power of deep
  learning: A tensor analysis,'' in \emph{Conference on learning theory}.\hskip
  1em plus 0.5em minus 0.4em\relax PMLR, 2016, pp. 698--728.

\bibitem{lizaire2024expressivityRNNsTensor}
M.~Lizaire, M.~Rizvi-Martel, M.~Gamal, and G.~Rabusseau, ``A tensor
  decomposition perspective on second-order {RNN}s,'' in \emph{Forty-first
  International Conference on Machine Learning}, 2024.

\bibitem{ali2023approximationTreeTensorNetworks}
M.~Ali and A.~Nouy, ``Approximation theory of tree tensor networks: Tensorized
  univariate functions,'' \emph{Constructive Approximation}, vol.~58, no.~2,
  pp. 463--544, 2023.

\bibitem{kileel2019expressivePolynomialNNs}
J.~Kileel, M.~Trager, and J.~Bruna, ``On the expressive power of deep
  polynomial neural networks,'' \emph{Adv. Neur. Inf. Proc. Syst.}, vol.~32,
  2019.

\bibitem{usevich2025identifiabilityDeepPNNs}
K.~Usevich, R.~Borsoi, C.~D{\'e}rand, and M.~Clausel, ``Identifiability of deep
  polynomial neural networks,'' \emph{Adv. Neur. Inf. Proc. Syst.}, 2025.

\bibitem{malgouyres2019multilinearCompressiveSensingLinearNNs}
F.~Malgouyres and J.~Landsberg, ``Multilinear compressive sensing and an
  application to convolutional linear networks,'' \emph{SIAM Journal on
  Mathematics of Data Science}, vol.~1, no.~3, pp. 446--475, 2019.

\bibitem{janzamin2015beatingPerilsTensorNeuralNets}
M.~Janzamin, H.~Sedghi, and A.~Anandkumar, ``Beating the perils of
  non-convexity: Guaranteed training of neural networks using tensor methods,''
  \emph{arXiv preprint arXiv:1506.08473}, 2015.

\bibitem{fornasier2021robustIdentificationShallowNeuralNets}
M.~Fornasier, J.~Vyb{\'\i}ral, and I.~Daubechies, ``Robust and resource
  efficient identification of shallow neural networks by fewest samples,''
  \emph{Information and Inference: A Journal of the IMA}, vol.~10, no.~2, pp.
  625--695, 2021.

\bibitem{dreesen2015decouplingPolinomials_1stOrder}
P.~Dreesen, M.~Ishteva, and J.~Schoukens, ``Decoupling multivariate polynomials
  using first-order information and tensor decompositions,'' \emph{SIAM J.
  Matrix Anal. Appl.}, vol.~36, no.~2, pp. 864--879, 2015.

\bibitem{ge2019learning2layerNNsSymmetricInputs}
R.~Ge, R.~Kuditipudi, Z.~Li, and X.~Wang, ``Learning two-layer neural networks
  with symmetric inputs,'' in \emph{International Conference on Learning
  Representations}, 2019.

\bibitem{chen2023learningPolynomialGenerativeModelsTensors}
S.~Chen, J.~Li, Y.~Li, and A.~R. Zhang, ``Learning polynomial transformations
  via generalized tensor decompositions,'' in \emph{Proceedings of the 55th
  Annual ACM Symposium on Theory of Computing}, 2023, pp. 1671--1684.

\bibitem{glasser2019expressiveTensorNetworksProbabilisticModeling}
I.~Glasser, R.~Sweke, N.~Pancotti, J.~Eisert, and I.~Cirac, ``Expressive power
  of tensor-network factorizations for probabilistic modeling,'' \emph{Adv.
  Neur. Inf. Proc. Syst.}, vol.~32, 2019.

\bibitem{mahajan2021tesseract}
A.~Mahajan, M.~Samvelyan, L.~Mao, V.~Makoviychuk, A.~Garg, J.~Kossaifi,
  S.~Whiteson, Y.~Zhu, and A.~Anandkumar, ``Tesseract: Tensorised actors for
  multi-agent reinforcement learning,'' in \emph{Proc. ICML}.\hskip 1em plus
  0.5em minus 0.4em\relax PMLR, 2021, pp. 7301--7312.

\bibitem{chen2022mixtureLinearClassifiersTensor}
A.~Chen, A.~De, and A.~Vijayaraghavan, ``Algorithms for learning a mixture of
  linear classifiers,'' in \emph{International Conference on Algorithmic
  Learning Theory}.\hskip 1em plus 0.5em minus 0.4em\relax PMLR, 2022, pp.
  205--226.

\bibitem{oneto2025ranks}
A.~Oneto and E.~Ventura, ``Ranks of tensors: geometry and applications,''
  \emph{Bollettino dell'Unione Matematica Italiana}, pp. 1--28, 2025.

\bibitem{comon2017identifiabilityXrank}
P.~Comon, Y.~Qi, and K.~Usevich, ``Identifiability of an {X-rank} decomposition
  of polynomial maps,'' \emph{SIAM Journal on Applied Algebra and Geometry},
  vol.~1, no.~1, pp. 388--414, 2017.

\bibitem{khrulkov2018expressiveTensorNNs}
V.~Khrulkov, A.~Novikov, and I.~Oseledets, ``Expressive power of recurrent
  neural networks,'' in \emph{ICLR}, 2018.

\bibitem{evert2022lra}
E.~Evert and L.~De~Lathauwer, ``Guarantees for existence of a best canonical
  polyadic approximation of a noisy low-rank tensor,'' \emph{SIAM J. Matrix
  Anal. Appl.}, vol.~43, no.~1, pp. 328--369, 2022.

\bibitem{arous2019landscape}
G.~B. Arous, S.~Mei, A.~Montanari, and M.~Nica, ``The landscape of the spiked
  tensor model,'' \emph{Communications on Pure and Applied Mathematics},
  vol.~72, no.~11, pp. 2282--2330, 2019.

\bibitem{mondelli2019connection2layerNnetTensorDec}
M.~Mondelli and A.~Montanari, ``On the connection between learning two-layer
  neural networks and tensor decomposition,'' in \emph{The 22nd International
  Conference on Artificial Intelligence and Statistics}.\hskip 1em plus 0.5em
  minus 0.4em\relax PMLR, 2019, pp. 1051--1060.

\bibitem{gilman2022tsvd}
K.~Gilman, D.~A. Tarzanagh, and L.~Balzano, ``Grassmannian optimization for
  online tensor completion and tracking with the t-svd,'' \emph{IEEE
  Transactions on Signal Processing}, vol.~70, pp. 2152--2167, 2022.

\bibitem{liu2023tensorCompressionNNsReview}
X.~Liu and K.~K. Parhi, ``Tensor decomposition for model reduction in neural
  networks: A review,'' \emph{IEEE Circuits and Systems Magazine}, vol.~23,
  no.~2, pp. 8--28, 2023.

\bibitem{zangrando2024geometryAwareTrainingTucker}
E.~Zangrando, S.~Schotth{\"o}fer, J.~Kusch, G.~Ceruti, and F.~Tudisco,
  ``Geometry-aware training of factorized layers in tensor {Tucker} format,''
  \emph{Proceedings, Adv. Neur. Inf. Proc. Syst.}, 2024.

\bibitem{yang2024loretta_lowRankTensorAdaptation}
Y.~Yang, J.~Zhou, N.~Wong, and Z.~Zhang, ``{LoRETTA}: Low-rank economic
  tensor-train adaptation for ultra-low-parameter fine-tuning of large language
  models,'' in \emph{Proc. Conference of the North American Chapter of the
  Association for Computational Linguistics: Human Language Technologies
  (Volume 1: Long Papers)}, 2024, pp. 3161--3176.

\bibitem{li2020generalizationNNsTensorLayersCompression}
J.~Li, Y.~Sun, J.~Su, T.~Suzuki, and F.~Huang, ``Understanding generalization
  in deep learning via tensor methods,'' in \emph{International Conference on
  Artificial Intelligence and Statistics}.\hskip 1em plus 0.5em minus
  0.4em\relax PMLR, 2020, pp. 504--515.

\bibitem{wang2024lowrankNN_tSVD_robustgeneralizationbias}
A.~Wang, C.~Li, M.~Bai, Z.~Jin, G.~Zhou, and Q.~Zhao, ``Transformed low-rank
  parameterization can help robust generalization for tensor neural networks,''
  \emph{Advances in Neural Information Processing Systems}, vol.~36, 2024.

\bibitem{papyan2020neuralCollapseNNpnas}
V.~Papyan, X.~Han, and D.~L. Donoho, ``Prevalence of neural collapse during the
  terminal phase of deep learning training,'' \emph{Proceedings of the National
  Academy of Sciences}, vol. 117, no.~40, pp. 24\,652--24\,663, 2020.

\bibitem{rendle2010factorizationMachines}
S.~Rendle, ``Factorization machines,'' in \emph{Proc. IEEE International
  conference on data mining}.\hskip 1em plus 0.5em minus 0.4em\relax IEEE,
  2010, pp. 995--1000.

\bibitem{levine2018deepQuantumEntanglementTensor}
Y.~Levine, D.~Yakira, N.~Cohen, and A.~Shashua, ``Deep learning and quantum
  entanglement: Fundamental connections with implications to network design,''
  in \emph{International Conference on Learning Representations}, 2018.

\bibitem{gribonval2022approximationDeepNNs}
R.~Gribonval, G.~Kutyniok, M.~Nielsen, and F.~Voigtlaender, ``Approximation
  spaces of deep neural networks,'' \emph{Constructive approximation}, vol.~55,
  no.~1, pp. 259--367, 2022.

\bibitem{cohen2016cnns_generalizedTensor_RelU}
N.~Cohen and A.~Shashua, ``Convolutional rectifier networks as generalized
  tensor decompositions,'' in \emph{International conference on machine
  learning}.\hskip 1em plus 0.5em minus 0.4em\relax PMLR, 2016, pp. 955--963.

\bibitem{awasthi2021efficient2LayerNNRelu}
P.~Awasthi, A.~Tang, and A.~Vijayaraghavan, ``Efficient algorithms for learning
  depth-2 neural networks with general {ReLU} activations,'' \emph{Adv. Neur.
  Inf. Proc. Syst.}, vol.~34, pp. 13\,485--13\,496, 2021.

\bibitem{oymak2021learningCNNsTensorsMoM}
S.~Oymak and M.~Soltanolkotabi, ``Learning a deep convolutional neural network
  via tensor decomposition,'' \emph{Information and Inference: A Journal of the
  IMA}, vol.~10, no.~3, pp. 1031--1071, 2021.

\bibitem{ge2018learning2layerNNsLandscapeDesign}
R.~Ge, J.~D. Lee, and T.~Ma, ``Learning one-hidden-layer neural networks with
  landscape design,'' in \emph{ICLR}, 2018.

\bibitem{dejonghe2023twolayer}
J.~De~Jonghe, K.~Usevich, P.~Dreesen, and M.~Ishteva, ``Compressing neural
  networks with two-layer decoupling,'' in \emph{Proc. 9th IEEE International
  Workshop on Computational Advances in Multi-Sensor Adaptive Processing
  (CAMSAP)}, 2023, pp. 226--230.

\bibitem{fornasier2022robustIdentification2layerNeuralNets}
M.~Fornasier, T.~Klock, and M.~Rauchensteiner, ``Robust and resource-efficient
  identification of two hidden layer neural networks,'' \emph{Constructive
  Approximation}, vol.~55, no.~1, pp. 475--536, 2022.

\end{thebibliography}
